\documentclass[11pt]{article}
\usepackage[utf8]{inputenc}
\usepackage[margin=1in]{geometry}
\usepackage{amsmath,amssymb,amsthm}
\usepackage{changepage}
\usepackage{textcomp}
\usepackage{cite}
\usepackage{nameref}
\usepackage{microtype}
\usepackage[table]{xcolor}
\usepackage{array}
\usepackage{graphicx}
\usepackage[labelfont=bf,labelsep=period,justification=raggedright,singlelinecheck=off]{caption}
\usepackage{enumitem}
\usepackage{float}
\usepackage{capt-of}
\usepackage{algorithm}
\usepackage{algpseudocode}
\usepackage{booktabs}
\usepackage{tikz}
\usetikzlibrary{shapes.geometric, arrows.meta, positioning, fit, backgrounds, calc}
\usepackage[colorlinks=true,linkcolor=blue!60!black,citecolor=blue!60!black,urlcolor=blue!60!black]{hyperref}
\newtheorem{assumption}{Assumption}
\newtheorem{proposition}{Proposition}
\newtheorem{corollary}{Corollary}

\makeatletter
\renewcommand{\@biblabel}[1]{#1.}
\makeatother

\title{\LARGE\bfseries Algorithmic drift in generative adaptive learning}
\author{
Firuz Kamalov\textsuperscript{1,$*$} \and Hana Sulieman\textsuperscript{2} \and Ahmed El Sayed\textsuperscript{1} \and Sivaraj R.\textsuperscript{3} \and Nitesh Sughnani\textsuperscript{4}\\[6pt]
{\small \textsuperscript{1}Department of Computational Sciences, Canadian University Dubai, Dubai, UAE}\\
{\small \textsuperscript{2}Department of Mathematics and Statistics, American University of Sharjah, Sharjah, UAE}\\
{\small \textsuperscript{3}Department of Mathematics, Dr. B.R. Ambedkar National Institute of Technology, Jalandhar, India}\\
{\small \textsuperscript{4}Knowledge and Human Development Authority, Dubai, UAE}\\
{\small \textsuperscript{$*$}Corresponding author: firuz@cud.ac.ae}
}
\date{}

\begin{document}
\maketitle

\begin{abstract}
\noindent Generative artificial intelligence promises to democratize personalized tutoring, yet the longitudinal stability of adaptive systems remains under-theorized. Current ethical discourse focuses on static bias. This paper argues that a distinct threat lies in \textit{algorithmic drift}: a feedback loop in which minor, demographically asymmetric observation errors compound over time into systemic inequities. We make three contributions. First, we formalize algorithmic drift as a coupled learner--tutor dynamical system, in which a knowledge-tracing estimator and a content policy operate in closed loop with the learner through a biased observation channel. In a reduced mean-field model we prove that an adaptive policy targeting a fixed success rate exhibits a bifurcation under an asymmetric false-negative channel. Below a critical noise threshold the protected group is merely served easier content; above it the learner's trajectory leaves the zone of proximal development and decays geometrically. Second, we corroborate this theory with an agent-based Monte Carlo simulation in which either a transparent online tracker or an LSTM deep-knowledge-tracing network drives an IRT-based curriculum policy. A matched-pair grading protocol executed on 3 open-weight large language models measures an asymmetric false-negative rate of $5$ percentage points for the most affected grader, the simulation baseline. Third, we introduce the \textit{temporal fairness stabilizer} (TFS), a targeted intervention that periodically re-probes learner capability, and benchmark it against standard exploration baselines with explicit group-fairness metrics and replication-level inference. Within the modeled environment, TFS closes the achievement gap where generic exploration does not, and we characterize the conditions under which it succeeds. We separate what is established analytically, what is demonstrated in simulation, and what remains to be validated with real learners. Our results reframe algorithmic fairness in education as a temporal property of a closed loop rather than a deployment-time checkbox.

\end{abstract}

\section{Introduction}

The integration of generative artificial intelligence (GenAI) into higher education signifies a fundamental shift from static content delivery to dynamic, responsive pedagogy. As educational institutions increasingly deploy AI-driven adaptive learning systems, the theoretical promise of a personalized curriculum tailored to the unique cognitive state of every learner appears increasingly viable. Recent scholarship suggests that these technologies can democratize access to high-quality tutoring, potentially advancing the United Nations SDG 4 by bridging resource disparities \cite{Kamalov2023}. However, while the computational efficiency and scalability of these systems are well-documented, their long-term systemic impact on equity remains under-theorized. As the field transitions from experimental pilots to institutional adoption, a critical question emerges regarding the longitudinal stability of these personalized interventions: does the optimization of individual learning pathways inadvertently automate the stratification of learners?

Current discourse on the ethics of AI in education predominantly focuses on static bias. This typically involves auditing training datasets for demographic representation or testing models for immediate discriminatory output, such as unfairly scoring an essay based on dialect. While necessary, this snapshot approach fails to capture the temporal and cumulative nature of the educational process. Learning is not a discrete event but a longitudinal trajectory defined by the sequential accumulation of knowledge. Consequently, the risks associated with AI adoption are not merely instantaneous errors but cumulative deviations. This paper posits that a significant threat to sustainable and equitable education lies in the phenomenon of \textit{algorithmic drift}. We define it as a feedback loop where minor, often imperceptible errors in an initial assessment compound over time, leading to significant divergences in learning outcomes between demographic groups.

This phenomenon is particularly acute in generative-adaptive systems which rely on deep knowledge tracing (DKT) to estimate learner proficiency. Unlike traditional platforms that route students through pre-determined logic trees, GenAI systems dynamically generate content based on a real-time estimation of the student's hidden knowledge state. This creates a bidirectional dependency: the model's prediction determines the content the student receives, and the student's performance on that content updates the model's future predictions. If a model misinterprets a student's struggle—confusing, for example, the linguistic friction of a non-native English speaker with a lack of conceptual understanding—it may erroneously classify the learner as low-proficiency. The system subsequently adapts by generating simplified remedial content. When the student successfully completes this easier material, the system’s pessimistic prediction is statistically reinforced. Over the duration of an academic term, this recursive dynamic can artificially suppress the trajectory of marginalized learners, inadvertently steering them into a lower-tier curriculum under the guise of personalization.

In this context, algorithmic bias evolves from a fairness issue into a sustainability issue. Instructional effort is misallocated not only when students drop out, but when capable students are systematically under-challenged and under-skilled by the infrastructure designed to support them. If AI-driven systems mistakenly predict failure for marginalized populations, they risk entrenching the very resource disparities they aim to solve. Achieving true sustainability in educational technology requires algorithms that are not just robust in static validation, but temporally fair over the lifecycle of a degree program.

To address this gap, our study presents a longitudinal audit of algorithmic amplification in adaptive learning environments. Concretely, we ask three research questions:
\begin{description}[leftmargin=1.6em, itemsep=2pt, topsep=3pt]
    \item[\textbf{RQ1.}] Under what conditions does an adaptive tutoring policy that is fair at initialization ($t=0$) diverge into demographically stratified outcomes as $t \to T$, and can this divergence be characterized analytically rather than merely observed empirically?
    \item[\textbf{RQ2.}] Do contemporary large language models, when deployed as zero-shot graders of free-text responses, exhibit the asymmetric false-negative behaviour that the model assumes, and at what magnitude?
    \item[\textbf{RQ3.}] Can a lightweight, fairness-aware regularization of the content policy interrupt the resulting feedback loop, and how does it compare to standard exploration strategies from reinforcement learning?
\end{description}

We answer these through a combination of dynamical-systems theory, simulation, and measurement. We first derive a reduced analytical model of the learner--tutor dyad and prove that asymmetric observation noise induces a bifurcation in the served-difficulty trajectory (RQ1). We then employ a discrete-time, agent-based Monte Carlo simulation of $10{,}000$ synthetic learner profiles over $50$ independent replications, introducing controlled observation noise to simulate linguistic barriers; the knowledge-tracing estimator is instantiated both as a transparent online tracker and as an LSTM deep-knowledge-tracing network run in the loop. We measure the noise parameter directly by executing a matched-pair grading protocol on 3 open-weight foundation models, which places the simulation baseline at the asymmetry measured for the most affected grader (RQ2). Finally, we propose and evaluate a regularizing intervention, the temporal fairness stabilizer (TFS), and benchmark it directly against $\epsilon$-greedy, upper-confidence-bound (UCB), and Thompson-sampling exploration, using explicit group-fairness metrics, to test whether generic exploration is sufficient or a fairness-targeted correction is required (RQ3). In summary, our contributions are fourfold. (i) We formalize algorithmic drift as a coupled learner--tutor dynamical system with an explicit trapping threshold. (ii) We execute and release a measurement of the demographically asymmetric grader noise that triggers drift. (iii) We propose the TFS intervention, evaluate it against RL baselines, analyze its failure conditions, and test its sensitivity to the learners' initial state and group heterogeneity. (iv) We propose a governance framework for the longitudinal auditing of adaptive learning systems.

The remainder of this paper is organized as follows. Section 2 reviews related work in intelligent tutoring systems and algorithmic fairness, emphasizing the limitations of static bias detection. Section 3 describes the methodology, detailing the simulation environment, the knowledge-tracing architectures, the formalization of algorithmic drift as a dynamical system (including a formal bifurcation result and its proof), and the formulation of the TFS together with the reinforcement-learning baselines against which it is compared. Section 4 presents the experimental results: the grader measurement, the algorithmic drift observed in the control scenario, the benchmark of interventions, the LSTM-DKT experiment, and sensitivity and robustness analyses. Finally, Section 5 separates what the evidence establishes from what remains to be validated, discusses the implications of these findings for educational sustainability, and provides recommendations for the governance of adaptive learning systems.
\section{Related Work}

The application of artificial intelligence in higher education (AIEd) has rapidly evolved from static computer-assisted instruction to dynamic intelligent tutoring systems (ITS) and, most recently, the integration of LLMs and GenAI \cite{kasneci2023chatgpt}. While traditional ITS relied on predefined pedagogical rules, GenAI introduces the capability to autonomously generate educational content, code explanations, and personalized feedback in real-time \cite{bommasani2021foundation, Kamalov2025b, Hultberg2024}. This transition aligns with the broader paradigm of ``precision education,'' which seeks to optimize learning outcomes by dynamically adapting curriculum difficulty to the individual learner’s zone of proximal development \cite{Kamalov2025a, williamson2020historical, Santos2026}.

Central to these adaptive environments is the task of knowledge tracing, which involves modeling a student’s changing knowledge state over time based on their interaction history. Since Bayesian knowledge tracing \cite{corbett1994knowledge}, adaptive tutors have been built from two modules: a knowledge-tracing estimator that infers a hidden mastery state from graded responses, and an instructional policy that selects the next activity from that estimate. In operation, these two modules and the learner form a closed loop. Deep knowledge tracing, introduced by Piech et al. \cite{piech2015deep}, advanced the field by utilizing recurrent neural networks and long short-term memory (LSTM) units to capture complex, non-linear dependencies in learning trajectories that traditional Bayesian methods often missed \cite{khajah2016deep}. A recent survey by Abdelrahman et al. \cite{abdelrahman2023knowledge} highlights that while DKT and its transformer-based successors have achieved state-of-the-art predictive performance, they function as ``black boxes,'' often prioritizing accuracy over interpretability and equitable calibration.

Despite the computational advancements, the AIEd community faces growing scrutiny regarding the sociotechnical implications of these systems. Extensive research has demonstrated that machine learning models frequently inherit sociological biases present in their training data \cite{chan2023}. In a comprehensive review, Baker and Hawn \cite{baker2021algorithmic} identify that educational algorithms often exhibit differential performance across demographic groups, particularly disadvantaging students from underrepresented racial backgrounds or those with non-standard linguistic patterns. In the context of natural language processing, Blodgett et al. \cite{blodgett2020language} argue that models trained on standard English corpora systematically penalize speakers of dialects or varieties like African American English, framing linguistic differences as errors or deficits. Two recent findings make this concern concrete for foundation-model graders: LLM-based text detectors misclassify essays by non-native English writers at far higher rates than those of native writers \cite{liang2023gpt}, and large language models make covertly prejudiced decisions about people on the basis of dialect alone \cite{hofmann2024dialect}.

Current approaches to mitigating these risks predominantly rely on static fairness auditing. It typically involves checking a model’s decision boundary at a single snapshot in time to ensure statistical parity or equalized odds before deployment \cite{kizilcec2022algorithmic, mehrabi2021survey}. However, scholars argue that static metrics are insufficient for educational settings, where the interaction between student and algorithm is continuous and cumulative \cite{holstein2022equity}.

Recent theoretical work on feedback loops in machine learning suggests that ensuring fairness at the initial time step ($t=0$) does not guarantee fairness over a long-term horizon ($t=n$). Hashimoto et al. \cite{hashimoto2018fairness} describe it as the risk of ``repeated risk minimization,'' where an algorithm’s minor errors in predicting the needs of a minority group lead to data shifts that reinforce the model’s initial error. The same closed-loop structure---a predictor whose outputs shape the data it is next evaluated on---underlies performative prediction \cite{perdomo2020performative} and the runaway feedback loops documented in predictive policing \cite{ensign2018runaway}, where the loop has been shown to operate in deployed systems rather than only in models of them. In the educational domain, this resembles the Matthew effect, where advantage begets advantage, and disadvantage compounds over time \cite{yu2023should}. Liu et al. \cite{liu2022equity} explicitly demonstrated this in knowledge tracing, showing that models optimizing for average accuracy can inadvertently generate inequitable curricula that stall the progress of specific learner groups. Porayska-Pomsta et al. \cite{porayska2024ethics} further argue that true sustainability in AIEd requires moving beyond simple efficiency metrics to consider the long-term human capital impact of these algorithmic feedback loops. 

In addressing sequential decision-making under uncertainty, RL and multi-armed bandit literature frequently employ exploration-exploitation paradigms (e.g., $\epsilon$-greedy, Upper Confidence Bound). However, standard exploration methods are designed to maximize average reward across an environment, not to rectify {asymmetric, demographic-specific observation noise}. While a standard exploration policy might briefly test a learner’s upper bounds, if structural linguistic friction persists, the system will rapidly retreat, interpreting the noisy failure as a true capability deficit. By synthesizing the literature on DKT with theories of longitudinal algorithmic dynamics, our study investigates how GenAI might inadvertently automate stratification, and we introduce a framework explicitly tailored to combat this temporal degradation rather than relying on generic exploration.

There has been increased focus on the transition towards highly dynamic AI-driven adaptive learning systems that utilize multimodal capabilities and LLMs to enhance educational outcomes \cite{Yuensook2026, Deckker2025}. A systematic review of agentic AI in education traces this transition from rule-based tutors to goal-oriented agents that perceive the learner, decide, and act in a closed loop \cite{Kamalov2026agentic}, which is precisely the loop whose longitudinal behaviour this paper analyzes. The integration of GenAI allows these platforms to move beyond static predefined rules by autonomously generating customized learning paths and tailoring instruction to meet the unique cognitive demands of students \cite{Kamaghe2026}. Furthermore, advanced intelligent tutoring systems are now employing automated multimodal knowledge graphs and retrieval-augmented generation to balance factual precision and explanatory richness \cite{Du2025, Noroozi2024}. 

However, the rapid deployment of these personalized systems amplifies concerns surrounding algorithmic fairness. As LLMs increasingly assume the role of intelligent tutors, recent evaluations reveal that these ``biased teachers'' can inadvertently generate and select educational content that perpetuates harmful stereotypes across demographic groups \cite{Weissburg2025}. Unchecked, AI-driven decision-making processes risk exacerbating existing educational disparities, cementing algorithmic bias as a compounding systemic barrier \cite{Boateng2025}. Recognizing this threat, scholars advocate for continuous algorithmic auditing and the incorporation of debiasing techniques directly into the model training pipeline \cite{Idowu2024}. Consequently, achieving equitable outcomes in AIEd necessitates rigorous longitudinal assessments and the implementation of targeted fairness interventions—such as the one proposed in this study—to mitigate discriminatory feedback loops before they permanently constrain a learner's educational trajectory.

\section{Methodology}

To systematically quantify the longitudinal impact of algorithmic bias in generative-adaptive learning systems, our study employs a discrete-time, agent-based Monte Carlo simulation. While we acknowledge that real-world educational ecosystems involve complex variables such as instructor intervention, peer support, and multi-modal feedback, a controlled computational simulation allows for the rigorous isolation of causal mechanisms. Specifically, our goal is to isolate the feedback loop between asymmetric observation noise and curriculum generation—a dynamic that is inherently confounded in empirical human-subject research. We model the educational environment as a partially observable Markov Decision process, where the adaptive tutoring system must estimate the hidden cognitive state of the learner based on noisy, sequential observations over time. The model is longitudinal at the \emph{individual} level: every synthetic learner carries its own latent ability, learning rate, estimated ability, served-difficulty history, and intervention state, and all population-level quantities reported below are aggregates of these individual trajectories across learners and across independent replications.

\subsection{Agent-Based Environment and Learning Dynamics}
\label{sec:env}
The simulation initializes a population $S$ of $N = 10,000$ synthetic learner agents interacting with the system over a longitudinal sequence of discrete time steps $t = 1, \dots, T$ (where $T=100$). The curriculum consists of $K=50$ distinct knowledge components, with difficulty parameters $\beta_j$ uniformly distributed over the unit interval $\beta_j \sim \mathcal{U}[0, 1]$. This ensures the curriculum spans the full spectrum of complexity, from remedial concepts to advanced mastery.

The probability of a student correctly answering a generated item is modeled using a standard item response theory (IRT) logistic function, which provides an interpretable abstraction of student-system interaction. The probability $P(c_{ij} = 1)$ that student $i$ answers item $j$ correctly at time $t$ is given by:
\begin{equation}
P(c_{ij} = 1) = \frac{1}{1 + e^{-a(\theta_i(t) - \beta_j)}}
\label{eq:irt}
\end{equation}
with a discrimination (slope) parameter $a=0.7$; setting $a=1$ recovers the pure Rasch link.

To address the oversimplification of simulated learning dynamics, we introduce substantial heterogeneity into the learner population. The population $S$ is stratified into a reference group ($S_{ref}$, $80\%$) representing learners with standard linguistic patterns and a protected group ($S_{prot}$, $20\%$) representing learners with non-standard linguistic markers. We emphasize that this binary stratification is a deliberate modeling abstraction: in practice, non-native speakers and speakers of non-standard dialects constitute heterogeneous populations that differ in the type and rate of grader friction they encounter, and the $20\%$ share is a stylized value chosen for exposition rather than a claim about any specific institution. Critically, the two groups are constructed to be \textit{ability-equivalent} at baseline---all agents are initialized with an identical latent-ability distribution $\theta_i(0) \sim \mathcal{N}(0.5, 0.1)$---so that any subsequent divergence is attributable to the observation channel rather than to prior differences. Agents further possess stochastic, heterogeneous learning rates $\alpha_i \sim \mathcal{N}(0.02, 0.005)$ (truncated to be positive). Ability equivalence is an identification device, not a claim about real cohorts. Section~\ref{sec:initial_state} relaxes it by varying the initial-state distribution and by giving the protected group a different prior ability or learning rate. The drift-attributable component of the gap is then isolated as the excess over the gap observed with a noiseless channel.

At each time step, the agent's true latent ability evolves based on the appropriateness of the pedagogical challenge. Let $\beta_t^{\text{srv}}$ denote the difficulty of the item actually served at time $t$ by the content policy (defined in Section~\ref{sec:arch}). Incorporating the individualized learning rate $\alpha_i$ and a cognitive forgetting factor $\lambda = 0.002$, the ability update is
\begin{equation}
    \theta_i(t{+}1) = \operatorname{clip}_{[0,1]}\!\Big(\theta_i(t) + \alpha_i \cdot g\big(\theta_i(t) - \beta_t^{\text{srv}}\big) - \lambda\,\theta_i(t)\Big),
    \label{eq:learn}
\end{equation}
where $g(\cdot)$ is the Zone-of-Proximal-Development (ZPD) gain and $\operatorname{clip}_{[0,1]}(x)=\min(\max(x,0),1)$ constrains ability to the same $[0,1]$ scale as item difficulty $\beta_j$. Our primary specification uses a smooth Gaussian ZPD kernel, $g(\Delta)=\exp\!\big(-\Delta^2/(2\omega^2)\big)$ with scale $\omega=0.12$, which yields full learning when the served difficulty matches ability and decays smoothly with mismatch; this graded form is what produces the smooth sensitivity profile reported in Section~\ref{sec:sensitivity_results} and cleanly separates systematic curriculum bias (which traps the protected group) from benign estimator jitter (which the reference group tolerates). For the theoretical analysis of Section~\ref{sec:theory} we use the hard-indicator idealization $g(\Delta)=\mathbb{I}(|\Delta|<\omega)$, which admits the closed-form trapping threshold of Proposition~\ref{prop:bifurcation}; the smooth kernel realizes the same threshold as a rapid-but-continuous transition. The clip is required for scale consistency: without it, the additive update has an interior equilibrium at $\theta^\star=\alpha_i g_{\max}/\lambda \gg 1$, off the difficulty scale of Eq.~\eqref{eq:irt}. Under the calibrated defaults, reference-group agents accumulate near the effective ceiling $\mu_{\text{final}}\approx 0.92$ (residual estimator jitter keeps them slightly below $1$).\footnote{Because the IRT link in Eq.~\eqref{eq:irt} uses $a(\theta-\beta)$ with $\theta,\beta\in[0,1]$, the effective range corresponds to success probabilities in $[\sigma(-a),\sigma(a)]$ at the extremes of the difficulty spectrum; we report $\theta$ throughout, and additionally report drift so that the estimator and ability remain commensurable.} To ensure robustness against this specific learning rule, we also evaluate the hard-indicator rule and a ``soft-gain'' variant in which learning occurs at a reduced rate $\alpha_i/4$ even when content falls outside the optimal ZPD. This comprehensive agent modeling ensures that any resulting longitudinal stratification is an artifact of the system's algorithmic mapping rather than of uniform-progression assumptions. Table~\ref{tab:params} summarizes the simulation parameters, their default values, and whether each is empirically motivated, calibrated, or assumed.

\begin{table}[H]
\begin{adjustwidth}{-0.5in}{-0.5in}
\centering
\scriptsize
\begin{tabular}{llll}
\toprule
\textbf{Symbol} & \textbf{Meaning} & \textbf{Default} & \textbf{Status} \\
\midrule
$N$, $T$, $K$ & learners, time steps, knowledge components & $10{,}000$, $100$, $50$ & design choice \\
$|S_{prot}|/N$ & protected-group share & $0.20$ & assumed (stylized) \\
$\theta_i(0)$ & initial latent ability & $\mathcal{N}(0.5,0.1)$, clipped & assumed; varied in Sec.~\ref{sec:initial_state} \\
$\alpha_i$ & learning rate & $\mathcal{N}(0.02,0.005)$, $>0$ & calibrated (reference ceiling $\approx0.92$) \\
$\lambda$ & forgetting rate & $0.002$ & assumed \\
$\omega$ & ZPD kernel width & $0.12$ & assumed; sets $\epsilon^\star$ \\
$a$ & IRT discrimination & $0.7$ & assumed; $a\in\{0.7,2.0\}$ in Sec.~\ref{sec:lstm_results} \\
$\beta_j$ & item difficulties & $\mathcal{U}[0,1]$, fixed bank & design choice \\
$\epsilon$ & false-negative rate, protected group & $0.05$ & empirically motivated (Sec.~\ref{sec:llm_results}); swept $0$--$0.30$ \\
$\eta$ & online-tracker step size & $0.2$ & calibrated; varied $0.1$--$0.4$ \\
$d$ & LSTM-DKT hidden size & $200$ & standard \cite{piech2015deep} \\
$(m,\tau,\delta)$ & TFS margin, patience, probe offset & $(0.10, 5, 0.10)$ & assumed; ablated in Sec.~\ref{sec:robustness} \\
$\epsilon_{expl}$, $c$, $\gamma$ & $\epsilon$-greedy, UCB, Thompson parameters & $0.1$, $0.3$, $0.9$ & standard defaults \\
$R$ & independent replications per condition & $50$ ($25$ for sweeps, $10$ for grids) & design choice \\
\bottomrule
\end{tabular}
\caption{Simulation parameters. ``Calibrated'' values were set so that the noiseless reference group reaches the documented mastery ceiling within the horizon; ``assumed'' values are modeling choices whose influence is examined in the sensitivity and robustness analyses.}
\label{tab:params}
\end{adjustwidth}
\end{table}

\subsection{Adaptive Architecture and Algorithmic Drift}
\label{sec:arch}
The automated tutoring system comprises two components: a knowledge-tracing inference engine and a generative content policy. Together with the learner and the observation channel, they form the coupled learner--tutor system studied in this paper. At each interaction $t$ the inference engine receives the assessed component $q_t$ and the binary observed correctness $a_t$, updates its belief about the learner, and emits an ability estimate $\hat\theta_{t+1}$ (equivalently, a mastery profile $\hat{p}_{t+1}\in[0,1]^K$ over the $K$ components). The generative content policy $\pi$ then targets the estimated ZPD by selecting a query $q_{t+1}$ whose difficulty tracks the estimate, $\beta_{t+1}=\hat\theta_{t+1}$ (the nearest item in the bank).

We instantiate the estimator in two forms. The \emph{online ability tracker} used for the full experimental sweeps is a stochastic-gradient maximum-likelihood update of the Rasch model,
\begin{equation}
    \hat\theta_{t+1} = \operatorname{clip}_{[0,1]}\!\Big(\hat\theta_t + \eta\big(a_{obs,t} - \sigma(a(\hat\theta_t - \beta_t))\big)\Big), \qquad \eta = 0.2,
    \label{eq:tracker}
\end{equation}
which raises the estimate after an observed success and lowers it after an observed failure, in proportion to how surprising the observation was. The \emph{LSTM deep-knowledge-tracing network} (hidden dimension $d=200$) \cite{piech2015deep} receives the input tuple $x_t=(\beta_t,a_{obs,t})$, updates its hidden state $h_t$, maps it to the predicted success probability on every item of the bank, and the ability estimate is read out as the Rasch ability whose success curve best fits this predicted profile. The network is pre-trained on noiseless synthetic interaction logs of a heterogeneous learner population generated by the same environment and is then run, frozen, inside the loop (Section~\ref{sec:lstm_results}). The first submission described the LSTM architecture but reported results only for the tracker; the revised Section~\ref{sec:lstm_results} reports the main experiment with the LSTM-DKT in the loop.

\paragraph{Why a learner--tutor loop with two tutor components.} The coupled system has three parts: the tutor's estimator, the tutor's policy, and the learner. They are connected in a cycle through the observation channel: the estimate sets the item difficulty, the item changes the learner's ability, and the graded answer feeds back into the estimate. The learner's ability and the tutor's estimate are the two coupled state variables. The estimator--policy decomposition of the tutor is not a simplification imposed for tractability. It is the canonical architecture of adaptive instruction. Every adaptive tutor since Bayesian knowledge tracing \cite{corbett1994knowledge} has been a controller of a partially observed process, and such a controller is, by construction, a state estimator composed with a decision rule. DKT-based and generative systems retain this structure. They replace the estimator by a neural network and the decision rule by a content generator \cite{piech2015deep,abdelrahman2023knowledge,Kamalov2025a,Kamalov2026agentic}. Additional modules of a deployed system fold into one of the two blocks or into the channels between them. The free-text grader is the observation channel. Hint generation, item authoring, and pacing rules are part of the policy. Instructor overrides and peer support are exogenous inputs to the learner. We do not claim that a deployed system consists of only these parts, nor that a particular product has been proved to behave this way. We claim that any closed-loop adaptive system contains this loop, so the mechanism analyzed here is a property it can exhibit. Whether it does so in a given deployment is an empirical question. Our grader measurement (Section~\ref{sec:llm_results}) begins to answer it, and Section~\ref{sec:evidence} states what has been established. Algorithmic drift is also \textit{not} an artifact of any particular estimator. As Section~\ref{sec:theory} shows, \emph{any} estimator that is driven by observed correctness and feeds a policy targeting a fixed success rate inherits a downward bias under an asymmetric false-negative channel. The mechanism therefore applies to Bayesian knowledge tracing, to the LSTM, and to transformer-based successors alike.

To empirically ground our findings in the era of GenAI, we augment this knowledge-tracing baseline with an LLM-Driven Knowledge Tracing (LLM-KT) measurement study. In modern agentic workflows, the correctness $a_t$ is not provided by a deterministic oracle but is inferred directly by prompting a foundational language model to evaluate free-text student responses. This introduces the vulnerability of linguistic friction. We formalize \textit{algorithmic drift} as the consequence of this observation noise. Let $a_{true}$ denote the student’s actual conceptual correctness, and $a_{obs}$ denote the response as graded by the AI. For the reference group, the observation is noiseless ($a_{obs} = a_{true}$). For the protected group, we introduce an asymmetric stochastic noise parameter $\epsilon$ representing false negatives triggered by linguistic friction:
\begin{equation}
P(a_{obs} = 0 \mid a_{true} = 1, s_i \in S_{prot}) = \epsilon
\label{eq:channel}
\end{equation}

By establishing this noise channel, algorithmic drift can be formalized as a recursive mapping in a dynamical system. The learner--tutor dyad functions as a coupled feedback loop. Given the asymmetric noise $\epsilon$, the system's updated belief $h_t = f(h_{t-1}, q_t, a_{obs})$ systematically diverges from the ideal noiseless state $h_t^*$. This depressed estimation $\hat{p}_t$ forces the generative policy to issue a simplified, remedial curriculum ($\beta_{t+1}$). Because the learner is consequently denied access to their true ZPD, their actual cognitive gain $\theta_t$ stalls, closing the loop and structurally validating the system's initial \emph{spurious deficit}---an estimated deficit that has no counterpart in true ability. The channel is deliberately minimal (a memoryless Bernoulli flip with a single rate); Section~\ref{sec:robustness} examines difficulty-dependent, temporally correlated, and continuously distributed per-learner friction, since real grader errors need not be constant or independent.

\subsection{Theoretical Analysis: A Bifurcation in the Learner--AI Dyad}
\label{sec:theory}
The simulation above is a high-fidelity but opaque instantiation of the feedback loop. To render the mechanism transparent---and to move from asserting that drift is a ``dynamical system'' to characterizing it---we analyze a reduced mean-field model that retains the two forces that matter: an adaptive policy that chases a target success rate, and an asymmetric observation channel. This yields a closed-form \emph{trapping threshold} that the full simulation then corroborates.

\begin{assumption}
\label{ass:reduced}
Consider a single mean-field knowledge component with true ability $\theta_t \in [0,1]$ and served difficulty $\beta_t$. The probability of a correct response is the Rasch link $s(\theta,\beta)=\sigma(\theta-\beta)$ with $\sigma(z)=(1+e^{-z})^{-1}$. The tutor observes correctness through a false-negative channel that flips $1\!\to\!0$ with probability $\epsilon_g$ ($\epsilon_{ref}=0$ for the reference group, $\epsilon_{prot}=\epsilon>0$ for the protected group), so the observed success rate is $s_{obs}(\theta,\beta)=(1-\epsilon_g)\,s(\theta,\beta)$. The policy is quasi-static (it adjusts $\beta$ faster than $\theta$ evolves) and sets $\beta_t$ so that the \emph{observed} success rate equals a fixed pedagogical target $p^\star \in (0,\sigma(\omega))$. Learning follows Eq.~\eqref{eq:learn}: $\theta$ increases while $|\theta-\beta|<\omega$ and otherwise decays at rate $\lambda$.
\end{assumption}

\begin{proposition}
\label{prop:bifurcation}
Under Assumption~\ref{ass:reduced}, define the served-difficulty offset $c_{\epsilon_g} := \sigma^{-1}\!\big(p^\star/(1-\epsilon_g)\big)$, so that the policy serves $\beta = \theta - c_{\epsilon_g}$, and define the critical noise level
\begin{equation}
    \epsilon^\star \;=\; 1 - \frac{p^\star}{\sigma(\omega)}.
    \label{eq:threshold}
\end{equation}
Then:
\begin{enumerate}[leftmargin=1.6em, itemsep=1pt, topsep=2pt]
    \item[(i)] For all $\epsilon \in (0,\epsilon^\star)$ the protected group is served strictly easier content than the reference group, with a difficulty deficit $\delta(\epsilon)=c_\epsilon-c_0 = \sigma^{-1}\!\big(p^\star/(1-\epsilon)\big)-\sigma^{-1}(p^\star)>0$ that is strictly increasing in $\epsilon$; yet both groups remain inside the ZPD and continue to learn.
    \item[(ii)] For $\epsilon \ge \epsilon^\star$ the protected group's served difficulty leaves the ZPD, the learning indicator collapses to $\mathbb{I}(\cdot)=0$, and ability decays geometrically as $\theta_{t+1}=(1-\lambda)\theta_t$, so the protected trajectory diverges monotonically from the reference trajectory.
    \item[(iii)] The asymptotic ability gap $\Delta_\infty(\epsilon)$ is non-decreasing in $\epsilon$ and is discontinuous at $\epsilon^\star$, exhibiting a bifurcation from a ``progress-but-under-challenged'' regime to a ``trapped-and-declining'' regime.
\end{enumerate}
\end{proposition}

\begin{proof}
The policy targets $s_{obs}(\theta,\beta)=p^\star$, i.e.\ $(1-\epsilon_g)\sigma(\theta-\beta)=p^\star$, giving $\theta-\beta = \sigma^{-1}\!\big(p^\star/(1-\epsilon_g)\big)=c_{\epsilon_g}$ and hence $\beta=\theta-c_{\epsilon_g}$. Because $\sigma^{-1}$ is strictly increasing and $p^\star/(1-\epsilon_g)$ is strictly increasing in $\epsilon_g$, the offset $c_{\epsilon_g}$ is strictly increasing in $\epsilon_g$; with $\epsilon_{ref}=0$ this yields $c_\epsilon>c_0$ and the strictly increasing deficit $\delta(\epsilon)$ in (i). Learning persists while $|\theta-\beta|=c_{\epsilon_g}<\omega$. Since $p^\star<\sigma(\omega)$ gives $c_0<\omega$, the reference group always learns. For the protected group, $c_\epsilon<\omega \iff p^\star/(1-\epsilon)<\sigma(\omega) \iff \epsilon<1-p^\star/\sigma(\omega)=\epsilon^\star$, establishing (i) for $\epsilon<\epsilon^\star$. At $\epsilon=\epsilon^\star$, $c_\epsilon=\omega$ and the indicator switches off; for $\epsilon\ge\epsilon^\star$, $c_\epsilon\ge\omega$ so $\mathbb{I}=0$ and Eq.~\eqref{eq:learn} reduces to $\theta_{t+1}=(1-\lambda)\theta_t$, proving the geometric decay in (ii). Finally, $\Delta_\infty(\epsilon)$ equals the reference ceiling minus the protected asymptote; on $(0,\epsilon^\star)$ the protected asymptote is non-increasing in $\epsilon$ because a larger offset $\delta(\epsilon)$ reduces effective time-in-ZPD, and for $\epsilon\ge\epsilon^\star$ the protected asymptote drops to the decay floor, giving the non-decreasing, discontinuous profile in (iii).
\end{proof}

Proposition~\ref{prop:bifurcation} makes three predictions that we test in Section~4. First, drift is present even at vanishing noise (part i), explaining why a $1\%$ false-negative rate already produces a measurable gap. Second, there exists a sharp onset of \emph{trajectory collapse} at $\epsilon^\star=1-p^\star/\sigma(\omega)$; for the parameters used here this predicts a transition in the low-double-digit noise range, consistent with the accelerating harm observed in our sensitivity analysis. Third, the harm is monotone in $\epsilon$. The result also clarifies the design target for any intervention: to prevent trapping, a policy must occasionally serve $\beta$ such that $|\theta-\beta|<\omega$ \emph{regardless} of the deflated observed signal---which is precisely the function of the TFS.

\begin{corollary}
\label{cor:tfs}
Suppose that every $\tau$ steps the policy overrides the observed-signal target and serves a probe difficulty $\beta^{\text{probe}}$ with $|\theta-\beta^{\text{probe}}|<\omega$. Then on probe steps $\mathbb{I}(\cdot)=1$ irrespective of $\epsilon$, so the protected ability satisfies $\theta_{t+\tau}\ge (1-\lambda)^{\tau-1}\theta_t + \alpha$, which is bounded below independently of the trapping condition in Proposition~\ref{prop:bifurcation}(ii). Hence periodic ZPD re-entry prevents unbounded geometric decay.
\end{corollary}

Corollary~\ref{cor:tfs} explains, at the level of the reduced model, \emph{why} a scheduled capability probe is sufficient to arrest drift, and---because the guarantee is conditional on the probe landing inside the ZPD---also why extreme noise that corrupts the probe itself can erode the TFS guarantee, a boundary regime we examine in simulation in Section~\ref{sec:robustness}.

Fig~\ref{fig:methodology_flow} illustrates the systemic architecture of our simulation, mapping algorithmic drift as a coupled dynamical system across three domains: the learner environment, the AI tutor system, and the observation channel.
As depicted, a learner's true cognitive state $(\theta_t)$ evolves based on curriculum difficulty $(\beta_t)$, generating a true correctness value $(a_{true})$. However, this signal must traverse a noisy observation channel. For marginalized groups, the LLM-KT Grader exhibits asymmetric linguistic friction $(\epsilon)$, introducing systemic false negatives into the system. This triggers the {algorithmic drift feedback loop} (highlighted on the left). The knowledge-tracing engine receives a depressed observed correctness $(a_{obs})$ and systematically underestimates learner mastery $(\hat{p}_{t+1})$. The generative policy subsequently issues remedial content below the learner's optimal ZPD. Consequently, the student's true cognitive growth stalls, cyclically reinforcing the AI's spurious deficit.
To interrupt the cycle, the TFS acts as a regulatory circuit breaker. By tracking a pessimism counter $(\kappa)$, TFS overrides the standard policy during prolonged remedial loops, forcefully issuing a high-difficulty item to re-evaluate the learner's true ceiling and decouple curriculum generation from the self-reinforcing observation noise.

\begin{center}
    \begin{tikzpicture}[
        >=Stealth,
        scale=0.7,
        transform shape,
        font=\sffamily,
        box/.style={rectangle, rounded corners=8pt, thick, draw=#1!80!black, fill=#1!5, minimum width=5.5cm, minimum height=2cm, align=center},
        intervention/.style={rectangle, rounded corners=8pt, densely dashed, very thick, draw=#1!80!black, fill=#1!5, minimum width=5.5cm, minimum height=1.8cm, align=center},
        groupbox/.style={rectangle, rounded corners=15pt, thick, draw=#1!50!black, fill=#1!2, inner sep=18pt},
        arrow/.style={->, thick, color=gray!80!black},
        boldarrow/.style={->, very thick, color=#1!90!black},
        tfsarrow/.style={->, very thick, densely dashed, color=green!70!black},
        label/.style={fill=white, inner sep=4pt, font=\small\sffamily, align=center}
    ]

        \node[box=blue] (dkt) at (-5, 2.5) {
            \textbf{Knowledge-Tracing Engine}\\[0.3ex]
            \footnotesize (online tracker / LSTM-DKT)\\[0.5ex]
            \footnotesize Updates belief: $h_t = f(h_{t-1}, q_t, a_{obs})$\\[-0.5ex] 
            \footnotesize Output: Predicted Mastery $\hat{p}_{t+1}$
        };
        
        \node[box=blue] (policy) at (5, 2.5) {
            \textbf{Generative Content Policy} $\pi(h_t)$\\[1ex] 
            \footnotesize Targets estimated ZPD\\[-0.5ex] 
            \footnotesize Output: Curriculum Difficulty $\beta_{t+1}$
        };
        
        \node[intervention=green] (tfs) at (5, 6) {
            \textbf{Temporal Fairness Stabilizer (TFS)}\\[1ex] 
            \footnotesize Tracks pessimism counter: $\kappa$\\[-0.5ex] 
            \footnotesize \textit{Action:} Triggers Forced Exploration Probe
        };

        \node[box=orange] (state) at (5, -2) {
            \textbf{Latent Cognitive State}\\[1ex] 
            \footnotesize $\theta_{t+1} = \theta_t + \alpha_i \cdot \mathbb{I}(|\theta_t - \beta_{t+1}| < \omega) - \lambda \theta_t$\\[-0.5ex] 
            \footnotesize (Heterogeneous Learning Rate $\alpha_i$)
        };
        
        \node[box=orange] (irt) at (5, -5.5) {
            \textbf{IRT Response Generator}\\[1ex] 
            \footnotesize $P(c_{ij}=1) = \left[1 + e^{-a(\theta_{t+1} - \beta_{t+1})}\right]^{-1}$\\[-0.5ex] 
            \footnotesize Output: True Correctness $a_{true}$
        };

        \node[box=red] (llm) at (-5, -5.5) {
            \textbf{LLM-KT Grader}\\[1ex] 
            \footnotesize \textit{Asymmetric Linguistic Friction}\\[-0.5ex] 
            \footnotesize $P(a_{obs}\!=\!0 \mid a_{true}\!=\!1, s_i \!\in\! S_{prot}) = \epsilon$
        };

        \begin{scope}[on background layer]
            \node[groupbox=blue, fit=(dkt) (policy) (tfs), label={[font=\bfseries\sffamily, text=blue!80!black]above:Adaptive Tutoring Architecture (AI System)}] (ai_group) {};
            
            \node[groupbox=orange, fit=(state) (irt), label={[font=\bfseries\sffamily, text=orange!80!black]below:Learner Environment}] (learner_group) {};
            
            \node[groupbox=red, fit=(llm), label={[font=\bfseries\sffamily, text=red!80!black]below:Observation Channel (Bias Injection)}] (llm_group) {};
        \end{scope}

        
        \draw[boldarrow=blue] (dkt) -- node[label, above] {Estimated Mastery\\$\hat{p}_{t+1}$} (policy);
        
        \draw[tfsarrow] (tfs) -- node[label, right] {Overrides if $\kappa \ge 5$\\(\textit{Interrupts Remedial Loop})} (policy);
        
        \draw[boldarrow=blue] (policy) -- node[label, right] {Generated Query $q_{t+1}$\\(Difficulty $\beta_{t+1}$)} (state);
        
        \draw[boldarrow=orange] (state) -- node[label, right] {True Ability\\$\theta_{t+1}$} (irt);
        
        \draw[boldarrow=orange] (irt) -- node[label, above] {True Correctness\\$a_{true}$} (llm);
        
        \draw[boldarrow=red] (llm) -- node[label, left, align=right] {
            \textbf{Observed Correctness $a_{obs}$}\\
            Systematic false negatives ($\epsilon$)\\
            \textit{Depresses future estimations}
        } (dkt);

        \node[font=\bfseries\sffamily, text=red!60!black, align=center] at (-1.2, -1.25) {ALGORITHMIC DRIFT\\[0.5ex]FEEDBACK LOOP};
        
        \draw[->, line width=4pt, color=red!20] (-1.2, 0.95) arc (90:-210:2.2cm);

    \end{tikzpicture}
    \captionof{figure}{Dynamical system representation of algorithmic drift in the adaptive learning environment. The recursive loop, illustrates how asymmetric observation noise ($\epsilon$) generated by linguistic friction depresses DKT mastery estimations ($\hat{p}_t$). This leads the generative policy to issue a remedial curriculum ($\beta_{t+1}$), which in turn stalls the learner's actual cognitive state ($\theta_{t+1}$). The TFS serves as a regulatory layer that interrupts this structural trajectory decoupling by forcing periodic exploration probes.}
    \label{fig:methodology_flow}
\end{center}

\subsection{Intervention: Temporal Fairness Stabilization (TFS)}
\label{sec:tfs}
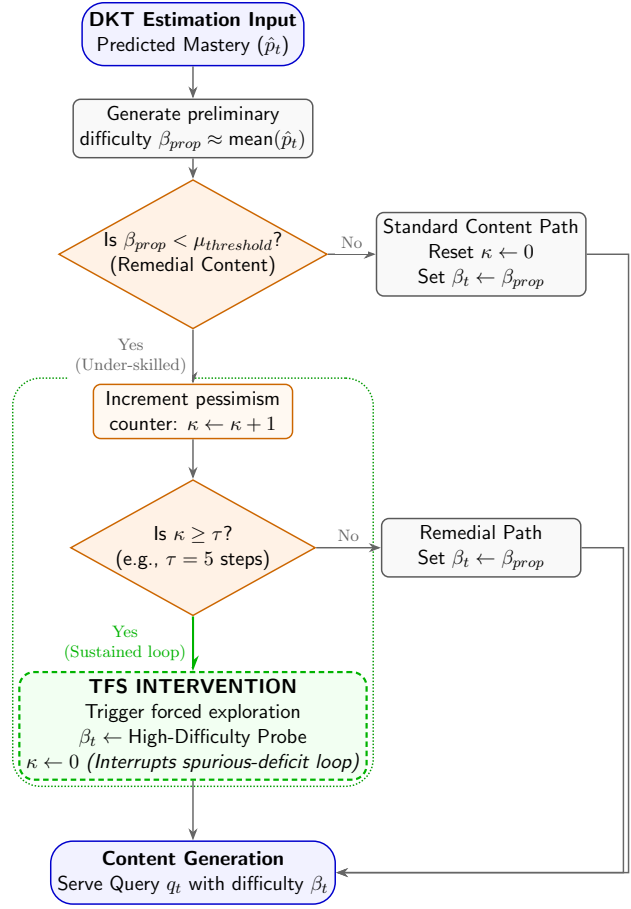
\begin{figure}[H]
    \centering
    \resizebox{0.5\textwidth}{!}{%
    \begin{tikzpicture}[
        >=Stealth,
        font=\sffamily,
        startend/.style={rectangle, rounded corners=12pt, thick, draw=blue!80!black, fill=blue!5, minimum width=4cm, minimum height=1.2cm, align=center},
        process/.style={rectangle, rounded corners=4pt, thick, draw=gray!70!black, fill=gray!5, minimum width=3.8cm, minimum height=1cm, align=center},
        decision/.style={diamond, aspect=1.8, thick, draw=orange!80!black, fill=orange!10, align=center, inner sep=1pt, minimum width=3.5cm},
        tfs_action/.style={rectangle, rounded corners=8pt, very thick, densely dashed, draw=green!70!black, fill=green!5, minimum width=5cm, minimum height=1.6cm, align=center},
        groupbox/.style={rectangle, rounded corners=15pt, thick, densely dotted, draw=green!60!black, fill=green!2, inner sep=15pt},
        arrow/.style={->, thick, color=gray!80!black},
        tfs_arrow/.style={->, very thick, color=green!70!black},
        label/.style={fill=white, inner sep=3pt, font=\footnotesize, align=center}
    ]

        \node[startend] (in) at (0, 0) {\textbf{DKT Estimation Input}\\Predicted Mastery ($\hat{p}_t$)};
        
        \node[process] (policy) at (0, -1.8) {Generate preliminary\\difficulty $\beta_{prop} \approx \text{mean}(\hat{p}_t)$};
        
        \node[decision] (check1) at (0, -4.2) {Is $\beta_{prop} < \mu_{threshold}$?\\(Remedial Content)};
        
        \node[process, draw=orange!80!black, fill=orange!5] (inc) at (0, -7.2) {Increment pessimism\\counter: $\kappa \leftarrow \kappa + 1$};
        
        \node[decision] (check2) at (0, -9.8) {Is $\kappa \ge \tau$?\\(e.g., $\tau=5$ steps)};
        
        \node[tfs_action] (tfs) at (0, -13.2) {\textbf{TFS INTERVENTION}\\Trigger forced exploration\\$\beta_t \leftarrow \text{High-Difficulty Probe}$\\$\kappa \leftarrow 0$ \textit{(Interrupts spurious-deficit loop)}};
        
        \node[startend] (out) at (0, -16) {\textbf{Content Generation}\\Serve Query $q_t$ with difficulty $\beta_t$};

        \node[process] (action1) at (5.5, -4.2) {Standard Content Path\\Reset $\kappa \leftarrow 0$\\Set $\beta_t \leftarrow \beta_{prop}$};
        
        \node[process] (action2) at (5.5, -9.8) {Remedial Path\\Set $\beta_t \leftarrow \beta_{prop}$};

        \begin{scope}[on background layer]
            \node[groupbox, fit=(inc) (check2) (tfs), label={[font=\bfseries\sffamily, text=green!60!black]above right:Temporal Fairness Stabilizer (Regularization Layer)}] (tfs_group) {};
        \end{scope}

        \draw[arrow] (in) -- (policy);
        \draw[arrow] (policy) -- (check1);
        
        \draw[arrow] (check1) -- node[label, above] {No} (action1);
        \draw[arrow] (check1) -- node[label, left] {Yes\\(Under-skilled)} (inc);
        
        \draw[arrow] (inc) -- (check2);
        
        \draw[arrow] (check2) -- node[label, above] {No} (action2);
        \draw[tfs_arrow] (check2) -- node[label, left] {Yes\\(Sustained loop)} (tfs);
        
        \draw[arrow] (tfs) -- (out);
        
        \draw[arrow] (action1.east) -- ++(0.8,0) |- (out.east);
        \draw[arrow] (action2.east) -- ++(0.8,0) |- (out.east);

    \end{tikzpicture}
    } 
    \caption{Flowchart of the TFS algorithm. Operating as a targeted regularization layer over the standard generative policy, TFS tracks a longitudinal pessimism counter ($\kappa$) to detect sustained periods of remedial curriculum assignment ($\beta < \mu_{threshold}$). Once the consecutive threshold ($\tau$) is reached, it overrides the depressed estimations of the DKT, forcing a diagnostic exploration probe to verify actual cognitive ceiling and break the algorithmic drift loop. For visual clarity the figure depicts the absolute-threshold form of the pessimism trigger; the deployed policy uses the cohort-relative criterion and capability ratchet of Algorithm~\ref{alg:tfs}.}
    \label{fig:tfs_flowchart}
\end{figure}

To interrupt this trajectory decoupling, we introduce the Temporal Fairness Stabilizer (TFS), a fairness-aware regularization layer integrated into the generative content policy (Algorithm~\ref{alg:tfs}; Fig~\ref{fig:tfs_flowchart} gives the corresponding flowchart). TFS tracks a longitudinal pessimism counter $\kappa$ for each student. A learner trapped by drift is characterized less by an \emph{absolute} difficulty floor than by being persistently held below the level the system assigns to comparable peers. The counter therefore increments whenever the learner's estimated ability sits more than a margin $m$ below the cohort's typical estimated level while the learner is not yet at the mastery ceiling. It resets on progress or after a probe. (An absolute variant, incrementing whenever $\beta<\mu_{threshold}=0.4$, is retained for the false-negative classification metric of Section~\ref{sec:metrics} and as an ablation option.) Once $\kappa$ reaches the patience parameter $\tau=5$, TFS issues a \emph{diagnostic} probe. Rather than a blind top-quartile item, it serves difficulty $\beta^{\text{probe}}=\hat\theta+\delta$, just above the possibly depressed estimate. This tests whether the learner is underestimated without overshooting the ZPD; a blind ceiling probe would fail uninformatively and leave the estimate uncorrected (cf.\ Corollary~\ref{cor:tfs}). Crucially, when a learner \emph{passes} a probe, the system ratchets its belief up to the demonstrated level ($\hat\theta\gets\max(\hat\theta,\beta^{\text{probe}})$). The ratchet resists the biased observation channel between probes. The hyperparameters $(m,\tau)$ are studied in a dedicated ablation (Section~\ref{sec:robustness}).

\begin{algorithm}[H]
\caption{Temporal Fairness Stabilizer (per learner, per step $t$)}
\label{alg:tfs}
\begin{algorithmic}[1]
\Require ability estimate $\hat\theta_t$, cohort estimates $\{\hat\theta_t^{(k)}\}$, counter $\kappa$, margin $m$, patience $\tau$, probe offset $\delta$, ceiling $c$
\State $\beta_{prop} \gets \hat\theta_t$ \Comment{policy's proposed difficulty (target ZPD)}
\State $q \gets \operatorname{Percentile}_{60}(\{\hat\theta_t^{(k)}\})$ \Comment{cohort reference level}
\If{$\hat\theta_t < q - m$ \textbf{and} $\hat\theta_t < c$} \Comment{held behind the cohort}
    \State $\kappa \gets \kappa + 1$
\Else
    \State $\kappa \gets 0$
\EndIf
\If{$\kappa \ge \tau$} \Comment{sustained pessimism detected}
    \State $\beta_t \gets \operatorname{clip}(\hat\theta_t + \delta)$ \Comment{diagnostic probe just above estimate}
    \State $\kappa \gets 0$
    \State \textbf{if} probe passed \textbf{then} $\hat\theta_t \gets \max(\hat\theta_t, \beta_t)$ \Comment{capability ratchet}
\Else
    \State $\beta_t \gets \beta_{prop}$
\EndIf
\State \Return served difficulty $\beta_t$
\end{algorithmic}
\end{algorithm}

It is crucial to distinguish TFS from generic exploration strategies utilized in reinforcement learning, such as $\epsilon$-greedy exploration or upper-confidence-bound (UCB) algorithms. Standard RL exploration techniques are designed to randomly or optimistically sample the action space to maximize global reward across an environment. However, they are fundamentally unequipped to rectify \textit{asymmetric, demographic-specific observation noise}. If a standard exploration policy probes a learner experiencing linguistic friction, the persistent observation noise will likely register the attempt as a failure, causing the agent to rapidly retreat and further cement its pessimistic estimate. Conversely, TFS is a targeted, deterministic fairness heuristic explicitly triggered by prolonged remedial assignment. It is designed not to maximize immediate reward, but to systematically break the spurious-deficit loop by enforcing periodic capability verification, ensuring that no learner is permanently relegated to a low-tier trajectory without reassessment.

\paragraph{Conditions under which TFS can be expected to work.} The guarantee of Corollary~\ref{cor:tfs} rests on four conditions. Each may fail in a deployment, so we state them explicitly. (C1) \emph{Probes are informative.} A probe is graded through the same channel as every other item, so with probability $\epsilon$ a passed probe is recorded as a failure. TFS tolerates this because the ratchet retains any demonstrated level and the next probe follows within $\tau$ steps. As $\epsilon$ grows, however, more probes are lost to the channel and the guarantee weakens (Section~\ref{sec:robustness}). (C2) \emph{A cohort reference exists.} The trigger of Algorithm~\ref{alg:tfs} compares a learner with the cohort's typical estimated level. In a very small cohort, or when the whole cohort is affected by friction, this relative trigger is uninformative and the absolute variant must be used instead (ablated in Section~\ref{sec:robustness}). (C3) \emph{Probes do not themselves harm the learner.} The model assumes that a failed probe leaves ability and engagement unchanged. In practice an unexpectedly difficult item may cause frustration, and a system that lowers its estimate after a failed probe entrenches the pessimism the probe was meant to test. Section~\ref{sec:probefail} quantifies both side effects. (C4) \emph{The probe reaches the ZPD.} The probe sits a fixed offset $\delta$ above the current estimate. If the estimate has fallen far below true ability, a single probe may be too easy to be diagnostic, and the ratchet lifts the estimate over successive probes rather than in one step.

\subsection{Exploration Baselines}
\label{sec:baselines}
To move this distinction from argument to evidence, we implement three standard exploration policies as comparators, all operating on the same DKT estimate and environment as TFS:
\begin{itemize}[leftmargin=1.4em, itemsep=1pt, topsep=2pt]
    \item {$\epsilon$-greedy:} with probability $\epsilon_{expl}=0.1$ the policy serves a difficulty drawn uniformly at random instead of $\operatorname{mean}(\hat{p}_t)$; otherwise it exploits.
    \item {UCB:} the policy serves the difficulty maximizing an optimism-augmented estimate $\hat{p}_t^{(k)} + c\sqrt{\ln t / n_k}$, where $n_k$ is the number of times component $k$ has been served and $c$ controls the exploration bonus.
    \item {Thompson sampling:} the policy maintains a Beta posterior over per-component mastery updated from observed correctness and serves difficulty proportional to a posterior sample, providing probability-matched exploration.
\end{itemize}
The decisive contrast is structural: $\epsilon$-greedy and UCB explore to reduce \emph{estimator} uncertainty and revert once the (noisy) signal is collected, whereas TFS explores to test a \emph{fairness} hypothesis---that a sustained remedial loop reflects grader friction rather than true deficit---and is therefore triggered by loop duration rather than by uncertainty. We report all four policies under identical noise conditions in Section~\ref{sec:baseline_results}.

\subsection{Evaluation Metrics and Implementation}
\label{sec:metrics}
To capture the full scope of algorithmic drift, we quantify system performance using three primary metrics that measure both internal system divergence and external learning outcomes. Because true ability $\theta$ and estimated mastery $\hat{p}$ must be compared on a common scale (see Section~\ref{sec:arch}), both are expressed in mastery-probability space before differencing:
\begin{itemize}
    \item {Drift divergence ($D$):} The mean absolute misalignment between the learner's true mastery and the system's estimated mastery, both in $[0,1]$. With the true mastery scalar $m_i(t)=\operatorname{clip}_{[0,1]}(\theta_i(t))$ and the estimate $\bar{p}_i(t)=\operatorname{mean}(\hat{p}_{i,t})$, we define
    \begin{equation}
        D = \frac{1}{T}\sum_{t=1}^{T}\big|\,m_i(t) - \bar{p}_i(t)\,\big|,
    \end{equation}
    which isolates the internal misalignment between truth and estimation and serves as a direct measure of epistemic degradation. (For a scalar difference the Euclidean and absolute forms coincide; we state the absolute form to avoid the impression of a vector norm.)
    \item {Gap widening ($\Delta$):} The difference in actual learning outcomes between the reference and protected groups, defined as $\Delta = (\bar{\theta}_{ref}(T) - \bar{\theta}_{ref}(0)) - (\bar{\theta}_{prot}(T) - \bar{\theta}_{prot}(0))$. This metric contextualizes the systemic harm of the drift relative to total potential learning gains.
    \item {False negative classification rate:} The proportion of temporal steps where an agent possesses high true capability ($\theta > 0.7$) but is erroneously served remedial content ($\beta < 0.4$) by the policy.
\end{itemize}
All three quantities are computed for both groups. Because the estimator lags behind a rising learner even without noise, $D$ is not zero for the reference group; the fairness-relevant quantity is the \emph{excess} of the protected group over the reference group, which the following metrics make explicit.

\paragraph{Group-fairness metrics.} To compare the two groups directly we report five metrics, each a protected-minus-reference (or ratio) contrast, averaged over $t=1,\dots,T$ and over replications:
\begin{itemize}[leftmargin=1.4em, itemsep=1pt, topsep=2pt]
    \item \emph{Exposure gap} $G_\beta = \bar\beta_{ref} - \bar\beta_{prot}$: the difference in mean served difficulty, a demographic-parity contrast on the system's treatment of the two groups.
    \item \emph{ZPD-access gap} $G_Z = \Pr(|\theta-\beta|<\omega \mid ref) - \Pr(|\theta-\beta|<\omega \mid prot)$: the difference in the share of steps on which the served item lies within the learner's ZPD, the quantity that drives learning in Eq.~\eqref{eq:learn}.
    \item \emph{Estimation-bias gap} $G_b = \overline{(\hat\theta-\theta)}_{prot} - \overline{(\hat\theta-\theta)}_{ref}$: the difference in signed calibration error of the estimator, a calibration-fairness contrast; a negative value means the protected group is underestimated relative to the reference group.
    \item \emph{Remedial-misclassification gap} $G_{FN} = \mathrm{FN}_{prot}-\mathrm{FN}_{ref}$: the difference in false-negative classification rates defined above, an equal-opportunity contrast (high-ability learners of each group should be served remedial content equally rarely).
    \item \emph{Gain ratio} $\rho = (\bar\theta_{prot}(T)-\bar\theta_{prot}(0))/(\bar\theta_{ref}(T)-\bar\theta_{ref}(0))$: the ratio of realized learning gains, a disparate-impact contrast; $\rho=1$ is parity and $\Delta$ above is the corresponding difference.
\end{itemize}

The simulation framework was developed in Python. The agent-based environment---IRT generation, the learner update, the noise channel, the knowledge-tracing estimator, and Monte Carlo sampling---is implemented in NumPy/SciPy and is fully vectorized across the $N=10{,}000$ agents (the only loop is over time), so the tracker-based experiment suite runs on a commodity CPU; the LSTM-DKT experiments use PyTorch on a single GPU. For the full sweeps the tutor's estimator is the transparent online tracker of Eq.~\eqref{eq:tracker}, because Proposition~\ref{prop:bifurcation} shows the drift mechanism is independent of the estimator class and the tracker maximizes interpretability and reproducibility; the main experiment is repeated with the LSTM-DKT in the loop in Section~\ref{sec:lstm_results}.

\paragraph{Ethics statement.} This study involved no human participants or personal data. All learners are synthetic agents, and the responses used in the grader measurement were written by the authors. No ethical approval or consent was required.

\paragraph{Statistical analysis.} The independent replication, not the simulated learner, is the unit of inference. Learners within a replication share the same curriculum bank, policy, and noise condition. They are therefore not independent observations of the condition's effect. Replications differ in every random draw: ability initialization, learning rates, group membership, response outcomes, and the noise channel. Every reported condition is therefore run over $R=50$ independent replications ($R=25$ for the one-dimensional sweeps and $R=10$ for two-dimensional grids). We report the mean across replications with a $95\%$ percentile-bootstrap confidence interval over replication means. Contrasts between policies are paired by seed: for each seed the same learner population is run under both policies. We report the mean paired difference, its bootstrap confidence interval, a paired $t$-test and Wilcoxon signed-rank test over the $R$ replications, and the fraction of replications in which the difference has the reported sign. Standardized effect sizes computed on replication means are not comparable to conventional benchmarks, because the between-replication standard deviation of a mean over thousands of learners is minuscule. We therefore report Cohen's $d$ at the learner level, pooled over replications, as a descriptive measure of the effect relative to between-learner variability. The two scales are labeled explicitly wherever both appear. \paragraph{Data availability.} All data underlying the findings are generated by the released code and are included with it. The repository \url{https://github.com/group-automorphism/algorithmic-drift} contains the simulator, the experiment scripts with their random seeds, every result file from which the tables and figures are built (\texttt{results/*.json}), the matched-pair item bank and the per-item grader verdicts of Section~\ref{sec:llm_results}, the pre-training and evaluation logs of the LSTM-DKT experiment, and the scripts that regenerate every number, table, and figure in the manuscript. No third-party or personal data were used.

\section{Results}
The results are presented in two phases. First, we execute a matched-pair grading protocol on current open-weight LLM graders to measure the observation-noise parameter $\epsilon$ and establish its order of magnitude. Second, we use our longitudinal Monte Carlo simulation to quantify the cumulative impact of this noise---specifically the divergence between the latent true knowledge state and the AI-estimated mastery across cohorts---with the online tracker and with the LSTM-DKT in the loop.

\subsection{Phase 1: Measuring Observation Noise in LLM Graders (RQ2)}
\label{sec:llm_results}

\paragraph{Protocol.} The protocol uses a matched-pair design. The item bank contains 40 conceptually correct free-text answers and 20 conceptually incorrect answers to short-answer prompts in introductory mechanics, energy, quantitative reasoning, and basic science, so that the effect is not tied to a single item. Within each pair, the reference response ($S_{ref}$) is written in Standard American English, while its counterpart ($S_{prot}$) is rewritten to carry second-language and non-standard markers (article omission, subject--verb agreement, tense, preposition calques, plural marking, word order) while holding the scientific content fixed. The bank was constructed by the authors and each pair was checked for conceptual equivalence before inclusion, so that linguistic form is the sole manipulated variable; the incorrect answers share the same two variants and are used to measure specificity (false positives). Each response is submitted to a grader operating as a zero-shot binary classifier under three instruction phrasings. All three direct the grader to judge conceptual mastery only and to ignore grammar, syntax, dialect, and fluency. Generation is greedy (temperature $0$) and the verdict is the modal verdict across the three phrasings. The grader's log-probabilities of the two admissible verdicts yield a calibrated $P(\mathrm{CORRECT})$ for each response. The statistics are the false-negative rate per variant on correct answers, their difference $\hat\epsilon=\mathrm{FN}_{prot}-\mathrm{FN}_{ref}$, an exact McNemar test on the paired verdicts, Cohen's $h$, the paired difference in $P(\mathrm{CORRECT})$ with a paired $t$-test, and the false-positive rate per variant on incorrect answers. Graders are 3 open-weight instruction-tuned models of the 4--7B class that are representative of the models deployed as local graders in tutoring pipelines (Table~\ref{tab:llm}), run locally in bfloat16 with versioned identifiers logged; the released code reproduces every number.

\begin{table}[H]
\begin{adjustwidth}{-0.5in}{-0.5in}
\centering
\scriptsize
\setlength{\tabcolsep}{2.5pt}
\begin{tabular}{lcccccccccc}
\toprule
\textbf{Grader} & $\mathrm{FN}_{ref}$ & $\mathrm{FN}_{prot}$ & $\hat\epsilon$ & McNemar $p$ & Cohen's $h$ & $\bar P(\mathrm{C})_{ref}$ & $\bar P(\mathrm{C})_{prot}$ & $\Delta P$ ($p$) & items $\downarrow$ & $\mathrm{FP}_{ref}$ / $\mathrm{FP}_{prot}$ \\
\midrule
Qwen2.5-7B-Instruct & 0.000 & 0.050 & $+0.050$ & 0.5 & $+0.45$ & 0.993 & 0.939 & $+0.055$ (0.043) & 95\% & 0.00 / 0.00 \\
Phi-4-mini-instruct & 0.000 & 0.000 & $+0.000$ & 1 & $+0.00$ & 0.995 & 0.993 & $+0.001$ (0.64) & 98\% & 0.10 / 0.05 \\
Mistral-7B-Instruct-v0.3 & 0.000 & 0.000 & $+0.000$ & 1 & $+0.00$ & 1.000 & 0.999 & $+0.001$ (0.31) & 62\% & 0.30 / 0.25 \\
\midrule
Pooled (3 graders) & & & $+0.017$ (mean) & 0.5 & & & & $+0.019$ (mean) & & \\
\bottomrule
\end{tabular}
\caption{Matched-pair LLM grader measurement (40 correct and 20 incorrect items, three instruction phrasings, modal verdicts). $\mathrm{FN}$: false-negative rate on conceptually correct answers; $\hat\epsilon=\mathrm{FN}_{prot}-\mathrm{FN}_{ref}$; McNemar: exact two-sided $p$ on the paired verdicts; $\bar P(\mathrm{C})$: mean grader probability of CORRECT; $\Delta P$: paired difference $\bar P(\mathrm{C})_{ref}-\bar P(\mathrm{C})_{prot}$ with its paired-$t$ $p$-value; items $\downarrow$: share of items on which $P(\mathrm{C})$ is lower for the second-language variant; $\mathrm{FP}$: false-positive rate on incorrect answers. The last row pools the discordant pairs over graders.}
\label{tab:llm}
\end{adjustwidth}
\end{table}

\paragraph{Findings.} Three findings emerge (Table~\ref{tab:llm}). First, the asymmetric channel exists and its direction is never reversed: every discordant pair across the three graders ($2$ of $2$) is a conceptually correct answer accepted in its Standard-English form and rejected in its second-language form, and no pair goes the other way. Second, its magnitude is grader-dependent. Qwen2.5-7B-Instruct rejects $5\%$ of the correct second-language answers that it accepts in Standard English ($\hat\epsilon=0.05$, Cohen's $h=+0.45$), while its calibrated probability of CORRECT is lower for the second-language variant on $95\%$ of items (paired difference $\Delta P=+0.055$, $p=0.043$); the asymmetry is therefore a systematic shift of the grader's confidence that crosses the decision boundary on a minority of items, which is precisely the memoryless false-negative channel of Eq.~\eqref{eq:channel}. Phi-4-mini-instruct and Mistral-7B-Instruct-v0.3 produce no hard false negatives on this bank, although their confidence is also lower for the second-language variant on 98\% and 62\% of items respectively, so the shift is present but sub-threshold for them. Third, the channel is asymmetric in the sense the model requires: none of the graders rejects correct Standard-English answers, and the false-positive rate on incorrect answers does not depend on the variant (Mistral-7B is simply lenient, accepting $30\%$ of incorrect answers in both variants). The measured rate for the most affected grader coincides with the simulation baseline $\epsilon=0.05$, and the sub-threshold shifts of the other graders show that the channel is a continuous property of grader confidence whose expression as a false-negative rate depends on the grader, the prompt, and the item.

\paragraph{What this measurement does and does not establish.} It establishes that the asymmetric false-negative channel assumed by the model exists in current graders, that it is grader-dependent, and that its magnitude for the most affected grader is $5$ percentage points---the simulation baseline $\epsilon=0.05$, which lies just above the trapping threshold $\epsilon^\star\approx0.040$ of the default calibration. It is a pilot measurement with a single, author-constructed dialect manipulation and a bank of 40 items. It therefore establishes existence and order of magnitude rather than a population value. With $40$ pairs the exact McNemar test on a $5$-point asymmetry is not significant ($p=0.5$), whereas the paired confidence shift is ($p=0.043$). A population estimate requires an independently annotated bank spanning dialects, subjects, and response lengths. It will also differ by grader, prompt, and version, as the variation across rows of Table~\ref{tab:llm} already shows; the released protocol is designed to be rerun as graders change. Independent of its measured value, the simulation treats $\epsilon$ as an exogenous parameter and characterizes outcomes across the full range $\epsilon\in[0,0.30]$ (Section~\ref{sec:sensitivity_results}). The conclusions can thus be read off for any grader whose asymmetry has been measured.

\subsection{Phase 2: Baseline Performance and Algorithmic Drift (Scenario A)}
Utilizing $\epsilon = 0.05$ as a baseline (a value within the range measured in Section~\ref{sec:llm_results}, and one below which the sensitivity analysis in Table~\ref{tab:sensitivity} shows the qualitative phenomenon persists), the control scenario demonstrated a failure to maintain epistemic accuracy for the protected group. While the reference group approached the mastery ceiling ($\mu_{final} = 0.922$), the protected group experienced distinct trajectory decoupling. All numbers in this and the following subsections are means over $R=50$ independent replications with replication-level $95\%$ confidence intervals (Section~\ref{sec:metrics}); they were regenerated from the released code for this revision and differ from the first submission in the third decimal.

\begin{figure}[H]
    \centering
    \includegraphics[width=\linewidth]{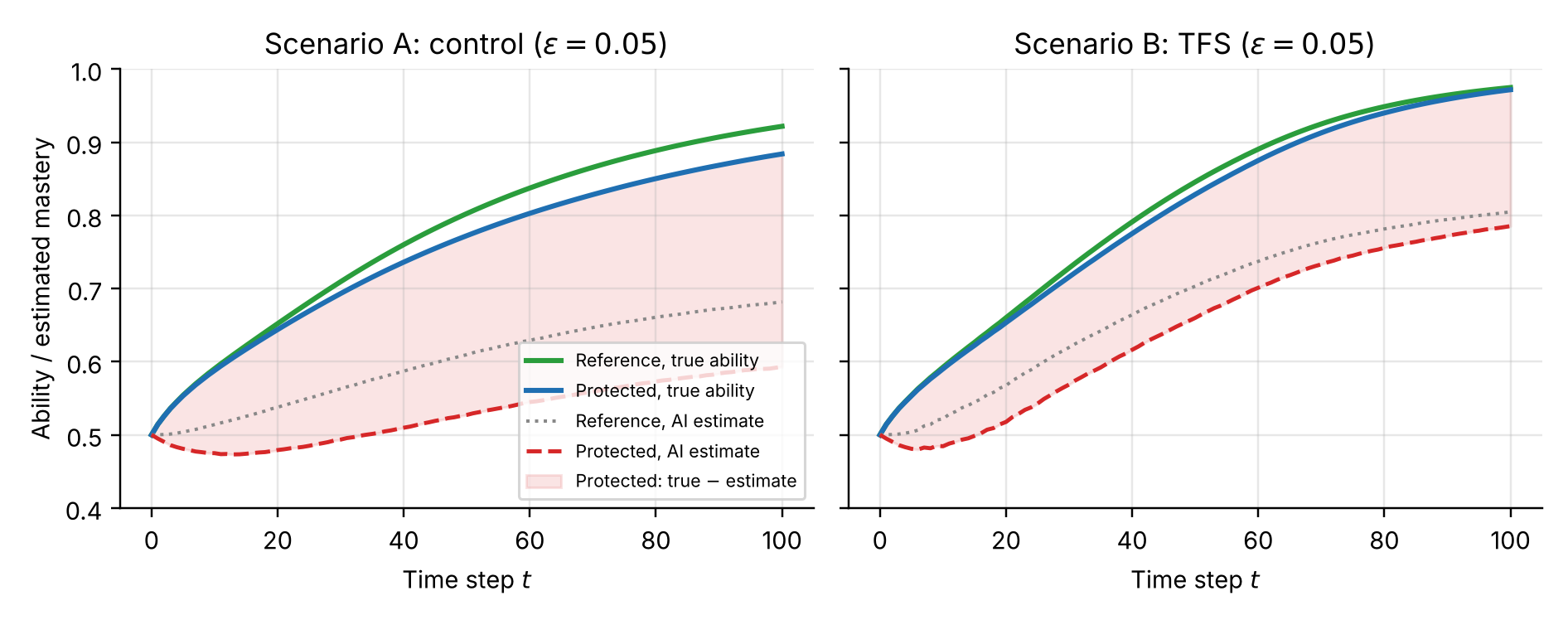} 
    \caption{{Algorithmic drift} ($\epsilon=0.05$, online tracker, mean trajectories over $R=50$ replications). Scenario A (left) illustrates the drift divergence, where the AI's estimation of the protected group (dashed) decouples from their true latent ability (blue solid line), suppressing their learning trajectory compared to the reference group (green). Scenario B (right) shows that the TFS mitigates this estimation error, substantially reducing the achievement gap under modeled conditions.}
    \label{fig:drift}
\end{figure}

Despite identical baseline ability distributions and the introduction of robust learner heterogeneity (including variable learning rates $\alpha_i$ and a cognitive forgetting factor $\lambda$), the protected group concluded with a mean latent ability of $\mu = 0.884$ ($95\%$ CI $[0.883, 0.885]$). The gap widening was $\Delta = 0.038$ ($95\%$ CI $[0.037, 0.039]$), meaning the protected group achieved roughly $9\%$ less growth than their peers at this noise level (gain ratio $\rho=0.911$); as Table~\ref{tab:sensitivity} shows, the gap widens substantially as $\epsilon$ increases past the trapping threshold (to $\Delta = 0.147$ at $\epsilon = 0.15$).

At this baseline noise level the \emph{capability} gap is moderate, but the \emph{epistemic} distortion is larger: the AI's internal belief state exhibited pronounced pessimism, with mean estimated mastery for the protected group at the end of the horizon of only $\hat{p} = 0.593$ against a true mean of $0.884$, whereas the reference group's estimate was $0.682$ against a true mean of $0.922$. Part of this underestimation is generic estimator lag, which affects both groups as their ability rises: the drift divergence is $D=0.268$ for the protected group against $D=0.236$ for the reference group, and the estimation-bias gap is $G_b=-0.047$ (reference group with a noiseless channel: $G_b=+0.000$). The protected-specific excess, visualized as the widening of the shaded region in Fig~\ref{fig:drift} (Scenario A) relative to the reference group, constitutes the algorithmic drift. The system attributes a spurious deficit to the protected group, suppressing curriculum difficulty (exposure gap $G_\beta=0.070$; ZPD-access gap $G_Z=0.059$) and keeping learners in a remedial loop that offers insufficient challenge to drive further growth. This decoupling---modest suppression of \emph{ability} but large corruption of the \emph{assessment record}---is the characteristic signature of drift just below the trapping threshold; as $\epsilon$ rises past $\epsilon^\star$, the ability suppression itself becomes large (Table~\ref{tab:sensitivity}). Because the simulation is deliberately isolated from external correctives (human instructors, multimodal assessment), these trajectories should be read as the behaviour of the autonomous algorithmic loop in isolation rather than as a forecast of any specific deployment.

\subsection{Impact of Temporal Fairness Stabilization (Scenario B)}
The TFS intervention substantially altered these dynamics. By forcing the generative policy to periodically issue diagnostic probe items triggered by sustained remedial assignment, and by retaining demonstrated capability through the ratchet, TFS corrected most of the model's internal drift. 

As shown in Fig~\ref{fig:drift} (Scenario B), the divergence between the reference and protected groups was substantially mitigated. Under these modeled conditions, the protected group achieved a final mean latent ability of $\mu = 0.972$ ($95\%$ CI $[0.971, 0.972]$), recovering essentially all of the lost learning potential relative to the control group. The gap widening was reduced to $\Delta = 0.003$ ($95\%$ CI $[0.002, 0.004]$; gain ratio $\rho=0.994$), and the drift divergence fell from $D=0.268$ to $D=0.203$ (estimation-bias gap from $G_b=-0.047$ to $-0.026$) as the periodic capability probes corrected the depressed estimate. The reference group is not harmed: because the cohort-relative trigger also probes reference learners who fall behind the cohort, the reference group's final ability rises from $0.922$ to $0.975$ (paired difference $+0.0530$, $95\%$ CI $[+0.0528, +0.0533]$).

While TFS appears remarkably effective in this controlled environment, we note that this near-complete recovery is partially an artifact of the stylized simulation design, where bypassing the noise channel immediately grants the system access to the student's true underlying cognitive state. In a real-world deployment, recovery would likely be noisier and more gradual due to compounding human factors, such as student disengagement or frustration accumulated during the preceding remedial loop. Nevertheless, the results suggest that targeted fairness corrections can systematically decouple curriculum generation from compounding observation noise.

\subsection{Comparison Against Exploration Baselines (RQ3)}
\label{sec:baseline_results}
A central claim of this paper is that generic RL exploration cannot substitute for a fairness-targeted correction. We test this directly by running the $\epsilon$-greedy, UCB, and Thompson-sampling policies of Section~\ref{sec:baselines} under the identical $\epsilon=0.05$ protected-group noise channel and comparing them to both the unmitigated control and TFS. Table~\ref{tab:baselines} reports the protected-group final ability, gap widening, and drift divergence for each policy (mean over $R=50$ replications, replication-level $95\%$ CI), together with the learner-level effect size against the control; Table~\ref{tab:fairness} reports the five group-fairness metrics of Section~\ref{sec:metrics} for the same runs.

\begin{table}[H]
\begin{adjustwidth}{-0.5in}{-0.5in}
\centering
\footnotesize
\begin{tabular}{lcccc}
\toprule
\textbf{Policy} & \textbf{Final ability ($S_{prot}$)} & \textbf{Gap $\Delta$} & \textbf{Drift $D$ ($S_{prot}$)} & \textbf{Learner-level $d$ vs.\ control} \\
\midrule
Control (no exploration) & $0.884$ [0.883, 0.885] & $0.038$ [0.037, 0.039] & $0.268$ & --- \\
$\epsilon$-greedy & $0.888$ [0.887, 0.889] & $0.035$ [0.034, 0.036] & $0.270$ & $+0.03$ \\
UCB & $0.901$ [0.900, 0.901] & $0.018$ [0.017, 0.019] & $0.283$ & $+0.13$ \\
Thompson sampling & $0.836$ [0.835, 0.837] & $0.019$ [0.019, 0.020] & $0.263$ & $-0.34$ \\
TFS (ours) & $\mathbf{0.972}$ [0.971, 0.972] & $\mathbf{0.003}$ [0.002, 0.004] & $\mathbf{0.203}$ & $+0.83$ \\
\bottomrule
\end{tabular}
\caption{Comparison of TFS against standard exploration baselines under $\epsilon=0.05$. Values are means over $R=50$ seeds with replication-level $95\%$ percentile-bootstrap CIs. The last column is Cohen's $d$ on protected-group final ability relative to the control policy, computed at the learner level and pooled over replications (positive favors the policy); the paired replication-level contrast for TFS is reported in Section~\ref{sec:stats}.}
\label{tab:baselines}
\end{adjustwidth}
\end{table}

\begin{table}[H]
\begin{adjustwidth}{-0.5in}{-0.5in}
\centering
\footnotesize
\begin{tabular}{lcccccc}
\toprule
\textbf{Policy} & $G_\beta$ & $G_Z$ & $G_b$ & $\mathrm{FN}_{prot}$ / $\mathrm{FN}_{ref}$ & $G_{FN}$ & $\rho$ \\
\midrule
Control, $\epsilon=0$ (parity reference) & $+0.000$ & $+0.000$ & $+0.000$ & $0.105$ / $0.105$ & $-0.001$ & $1.000$ \\
Control (no exploration) & $+0.070$ & $+0.059$ & $-0.047$ & $0.138$ / $0.105$ & $+0.033$ & $0.911$ \\
$\epsilon$-greedy & $+0.064$ & $+0.052$ & $-0.049$ & $0.145$ / $0.115$ & $+0.030$ & $0.916$ \\
UCB & $+0.059$ & $+0.046$ & $-0.056$ & $0.098$ / $0.066$ & $+0.032$ & $0.957$ \\
Thompson sampling & $+0.065$ & $+0.022$ & $-0.058$ & $0.127$ / $0.105$ & $+0.022$ & $0.945$ \\
TFS (ours) & $+0.033$ & $+0.045$ & $-0.026$ & $0.037$ / $0.032$ & $+0.006$ & $0.994$ \\
\bottomrule
\end{tabular}
\caption{Group-fairness metrics at $\epsilon=0.05$ (means over $R=50$ replications; time-averaged over $t=1,\dots,T$). $G_\beta$: exposure gap (reference minus protected mean served difficulty); $G_Z$: ZPD-access gap; $G_b$: estimation-bias gap (protected minus reference signed calibration error; negative = protected group underestimated relative to reference); $\mathrm{FN}_{prot}/\mathrm{FN}_{ref}$: remedial-misclassification rates with their gap $G_{FN}$; $\rho$: gain ratio. The first row is the control policy with a noiseless channel ($\epsilon=0$), the parity reference point. Replication-level $95\%$ CI half-widths are at most $0.002$ for every entry.}
\label{tab:fairness}
\end{adjustwidth}
\end{table}

The measured ordering in Tables~\ref{tab:baselines} and~\ref{tab:fairness} is consistent with the qualitative prediction of Proposition~\ref{prop:bifurcation}. With a noiseless channel every fairness contrast is at parity (first row of Table~\ref{tab:fairness}), which confirms that the metrics are not biased by the stochastic estimator itself. Under the biased channel, $\epsilon$-greedy explores around the \emph{deflated} observed signal and then reverts to exploiting it, so it leaves the gap essentially intact ($\Delta=0.035$ versus the control's $0.038$). Thompson sampling, whose probability-matched shifts are driven by the recent observed success rate, reduces the \emph{difference} between groups ($\Delta=0.019$) only by lowering the served difficulty for \emph{both} groups: the reference group's final ability falls from $0.922$ to $0.856$ and the protected group's to $0.836$, so the apparent fairness gain is purchased with a loss of learning for everyone. UCB, whose optimism systematically probes above the estimate, helps partially ($\Delta=0.018$) but does not close the gap and raises the drift divergence, because its optimism inflates the estimate without correcting it. Only TFS---which conditions on loop \emph{duration} rather than uncertainty, re-enters the ZPD regardless of the deflated signal, and ratchets the estimate to demonstrated capability (Corollary~\ref{cor:tfs})---brings every contrast close to parity ($\Delta=0.003$, $G_{FN}=+0.006$, $\rho=0.994$) while raising, not lowering, the reference group's outcome. These results support the paper's central claim within the modeled assumptions: generic exploration is not a substitute for a fairness-targeted correction.

\subsection{Main Experiment with an LSTM-DKT Estimator in the Loop}
\label{sec:lstm_results}
\begin{figure}[H]
    \centering
    \includegraphics[width=\linewidth]{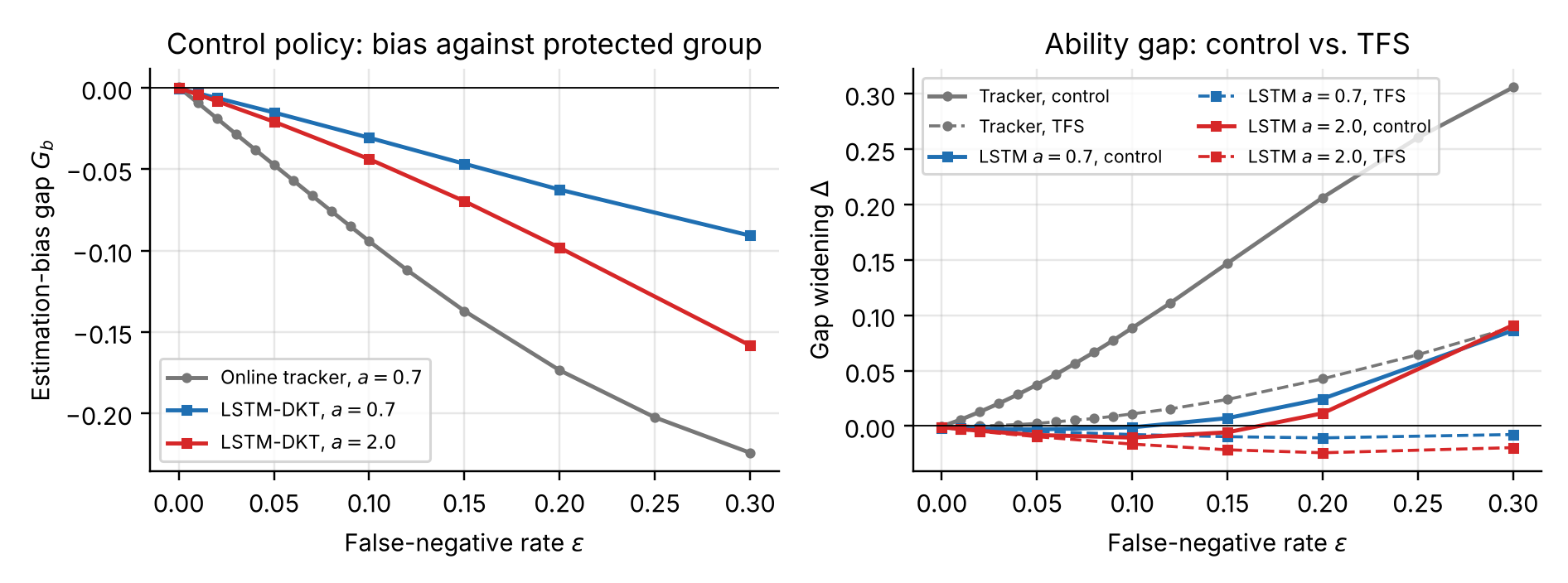}
    \caption{Online tracker versus LSTM-DKT in the loop. Left: estimation-bias gap $G_b$ under the control policy as a function of $\epsilon$ (negative: protected group under-estimated relative to the reference group). Right: ability gap $\Delta$ under control (solid) and TFS (dashed). The LSTM reproduces the estimation bias at every $\epsilon$ and the trapping transition at larger $\epsilon$; TFS holds the gap at or below zero for both estimators.}
    \label{fig:lstm}
\end{figure}

The results above use the online tracker of Eq.~\eqref{eq:tracker}. To verify that the mechanism, the gap, and the intervention's effect are not artifacts of this estimator, we repeated the main experiment with the LSTM deep-knowledge-tracing network of Section~\ref{sec:arch} as the in-the-loop estimator. The network (hidden size $200$) was pre-trained for 8 epochs on 24,000 noiseless synthetic interaction logs of $100$ steps, generated by the same environment for a \emph{heterogeneous} population with initial ability drawn uniformly on $[0.05,0.95]$, with the standard DKT objective of predicting the next observed outcome. It was then frozen and run inside the loop with one hidden state per learner. The ability estimate is read out from its predicted mastery profile over the item bank. A network trained instead on the narrow deployment prior $\mathcal{N}(0.5,0.1)$ learns item difficulty but collapses to the population prior over learners, because each binary verdict carries too little information at low discrimination to separate learners whose abilities differ by $0.1$. Such a network is unresponsive to observations. It exhibits neither tracking nor drift, which is consistent with the theory (drift requires an estimator that responds to the observed signal) but uninformative as a test. The heterogeneous pre-training population is also the realistic case, since a deployed knowledge-tracing model is trained on logs from many cohorts. The information content of a binary outcome is governed by the IRT discrimination $a$, so we ran the experiment at $a\in\{0.7,1.0,2.0,3.0\}$. Table~\ref{tab:lstm} reports the results for $a=0.7$ and $a=2.0$: the held-out AUC of the network against the \emph{oracle} AUC of the true success probability (the ceiling imposed by the link), and the gap and fairness metrics for control and TFS over the noise grid, with $R=20$ replications per condition. The tracker's gaps are listed alongside for comparison. Fig~\ref{fig:lstm} plots the estimation-bias gap and the ability gap against $\epsilon$ for both estimators.

\begin{table}[H]
\begin{adjustwidth}{-0.5in}{-0.5in}
\centering
\footnotesize
\begin{tabular}{lcccccc}
\toprule
\multicolumn{7}{l}{\emph{$a=0.7$: LSTM-DKT held-out AUC $0.553$ (heterogeneous) / $0.553$ (deployment); oracle AUC $0.565$ / $0.559$}} \\
$\epsilon$ & Control $\Delta$ & Control $G_b$ & Control $G_\beta$ & TFS $\Delta$ & TFS $G_b$ & Tracker $\Delta$: control / TFS \\ \midrule
0.00 & $-0.002$ [-0.003, -0.000] & $-0.000$ & $+0.000$ & $-0.002$ [-0.003, -0.000] & $-0.000$ & $+0.001$ / $+0.001$ \\
0.02 & $-0.002$ [-0.004, -0.001] & $-0.006$ & $+0.008$ & $-0.003$ [-0.004, -0.002] & $-0.005$ & $+0.014$ / $+0.001$ \\
0.05 & $-0.003$ [-0.004, -0.002] & $-0.015$ & $+0.022$ & $-0.005$ [-0.006, -0.004] & $-0.013$ & $+0.039$ / $+0.003$ \\
0.10 & $-0.001$ [-0.002, -0.000] & $-0.031$ & $+0.047$ & $-0.008$ [-0.009, -0.007] & $-0.026$ & $+0.089$ / $+0.012$ \\
0.15 & $+0.007$ [0.006, 0.008] & $-0.047$ & $+0.075$ & $-0.010$ [-0.011, -0.009] & $-0.038$ & $+0.148$ / $+0.025$ \\
0.20 & $+0.025$ [0.024, 0.026] & $-0.063$ & $+0.106$ & $-0.011$ [-0.012, -0.010] & $-0.050$ & $+0.206$ / $+0.043$ \\
0.30 & $+0.086$ [0.085, 0.088] & $-0.091$ & $+0.169$ & $-0.008$ [-0.009, -0.007] & $-0.070$ & $+0.306$ / $+0.090$ \\
\midrule
\multicolumn{7}{l}{\emph{$a=2.0$: LSTM-DKT held-out AUC $0.640$ (heterogeneous) / $0.635$ (deployment); oracle AUC $0.658$ / $0.648$}} \\
$\epsilon$ & Control $\Delta$ & Control $G_b$ & Control $G_\beta$ & TFS $\Delta$ & TFS $G_b$ & Tracker $\Delta$: control / TFS \\ \midrule
0.00 & $-0.001$ [-0.002, 0.000] & $-0.000$ & $-0.000$ & $-0.001$ [-0.002, -0.000] & $-0.000$ & $+0.000$ / $+0.001$ \\
0.02 & $-0.004$ [-0.005, -0.003] & $-0.008$ & $+0.008$ & $-0.005$ [-0.006, -0.004] & $-0.007$ & $+0.004$ / $+0.000$ \\
0.05 & $-0.008$ [-0.009, -0.007] & $-0.021$ & $+0.023$ & $-0.010$ [-0.011, -0.008] & $-0.018$ & $+0.013$ / $-0.000$ \\
0.10 & $-0.010$ [-0.011, -0.009] & $-0.044$ & $+0.053$ & $-0.016$ [-0.018, -0.015] & $-0.037$ & $+0.035$ / $+0.001$ \\
0.15 & $-0.006$ [-0.007, -0.005] & $-0.070$ & $+0.093$ & $-0.022$ [-0.023, -0.020] & $-0.057$ & $+0.074$ / $+0.005$ \\
0.20 & $+0.011$ [0.011, 0.012] & $-0.098$ & $+0.141$ & $-0.024$ [-0.025, -0.023] & $-0.079$ & $+0.129$ / $+0.011$ \\
0.30 & $+0.091$ [0.090, 0.092] & $-0.158$ & $+0.258$ & $-0.020$ [-0.021, -0.019] & $-0.124$ & $+0.262$ / $+0.038$ \\
\bottomrule
\end{tabular}
\caption{Main experiment with the LSTM-DKT in the loop ($N=10{,}000$, $T=100$, $R=20$ replications per cell; replication-level $95\%$ CIs in brackets). Held-out next-outcome AUC of the network on the heterogeneous pre-training population and on the deployment population, against the oracle AUC of the true success probability. $\Delta$: gap widening; $G_b$: estimation-bias gap; $G_\beta$: exposure gap. The last column gives the online tracker's gap at the same $(a,\epsilon)$ ($R=10$).}
\label{tab:lstm}
\end{adjustwidth}
\end{table}

Four findings emerge (Table~\ref{tab:lstm}, Fig~\ref{fig:lstm}). First, the network is a competent and well-calibrated estimator: its held-out AUC is within $0.02$ of the oracle ceiling at every discrimination level, with a noiseless channel both groups reach $0.979$ by the end of the horizon (against $0.922$ with the tracker), the drift divergence is $D=0.126$ (against $0.236$), and every fairness contrast is at parity. Second, the estimation and exposure links of the mechanism are reproduced at every noise level and every discrimination: under the biased channel the protected group is under-estimated relative to the reference group and served easier content, monotonically in $\epsilon$ (at $a=0.7$, $G_b$ from $-0.015$ and $G_\beta=+0.022$ at $\epsilon=0.05$ to $G_b=-0.091$ and $G_\beta=+0.169$ at $\epsilon=0.30$; at $a=2.0$, $G_b=-0.021$ and $G_\beta=+0.023$ at $\epsilon=0.05$). The spurious deficit and the remedial exposure that the theory predicts are therefore properties of the loop, not of the tracker. Third, the better-calibrated estimator widens the exposure-drift regime of Proposition~\ref{prop:bifurcation}(i) before trapping sets in: because the LSTM keeps the served difficulty inside the ZPD over a wider range of offsets, the ability gap remains at parity up to $\epsilon\approx0.10$ (it is even marginally negative: for $a\ge1$ the network is slightly optimistic---estimate $0.988$ against true ability $0.973$ at $\epsilon=0$, $a=2.0$---and a small false-negative rate partly offsets this optimism for the protected group) and turns positive at $\epsilon\approx0.15$ for $a=0.7$ and $0.20$ for $a=2.0$, reaching $\Delta=+0.025$ at $\epsilon=0.20$ and $+0.086$ at $\epsilon=0.30$ ($a=0.7$; $+0.091$ at $a=2.0$), with the same accelerating profile as the tracker. This is the trapping threshold of the theory expressed through the estimator's effective offset: a well-calibrated estimator tolerates more grader noise before learning is harmed, but the corruption of the assessment record and of content exposure begins at the smallest $\epsilon$ regardless. Fourth, TFS is effective and safe with the learned estimator: it holds the protected group at or above the reference group at every $\epsilon$ and every $a$ (largest residual $\Delta=-0.001$; at $\epsilon=0.30$, $\Delta=-0.008$ against $+0.086$ under control for $a=0.7$ and $-0.020$ against $+0.091$ for $a=2.0$), with $1.2$--$4.7$ probes per learner and a reference-group outcome never below $0.964$. The exploration baselines again fail to substitute for it: $\epsilon$-greedy leaves the contrasts unchanged ($\Delta=-0.002$ at $\epsilon=0.05$), Thompson sampling lowers the protected group's ability to $0.896$ and widens the gap ($\Delta=+0.019$), and UCB's optimism bonus, added to an already well-calibrated estimate, pushes the served difficulty out of the ZPD for everyone (protected-group ability $0.736$ against $0.981$ under control). The results for $a=1.0$ and $a=3.0$ follow the same pattern (onset of the ability gap at $\epsilon\approx0.30$ for $a=3.0$) and are included in the released result files.

\subsection{Sensitivity Analysis: Impact of Noise Parameter ($\epsilon$)}
\label{sec:sensitivity_results}

\begin{figure}[h]
    \centering
     \includegraphics[width=\linewidth]{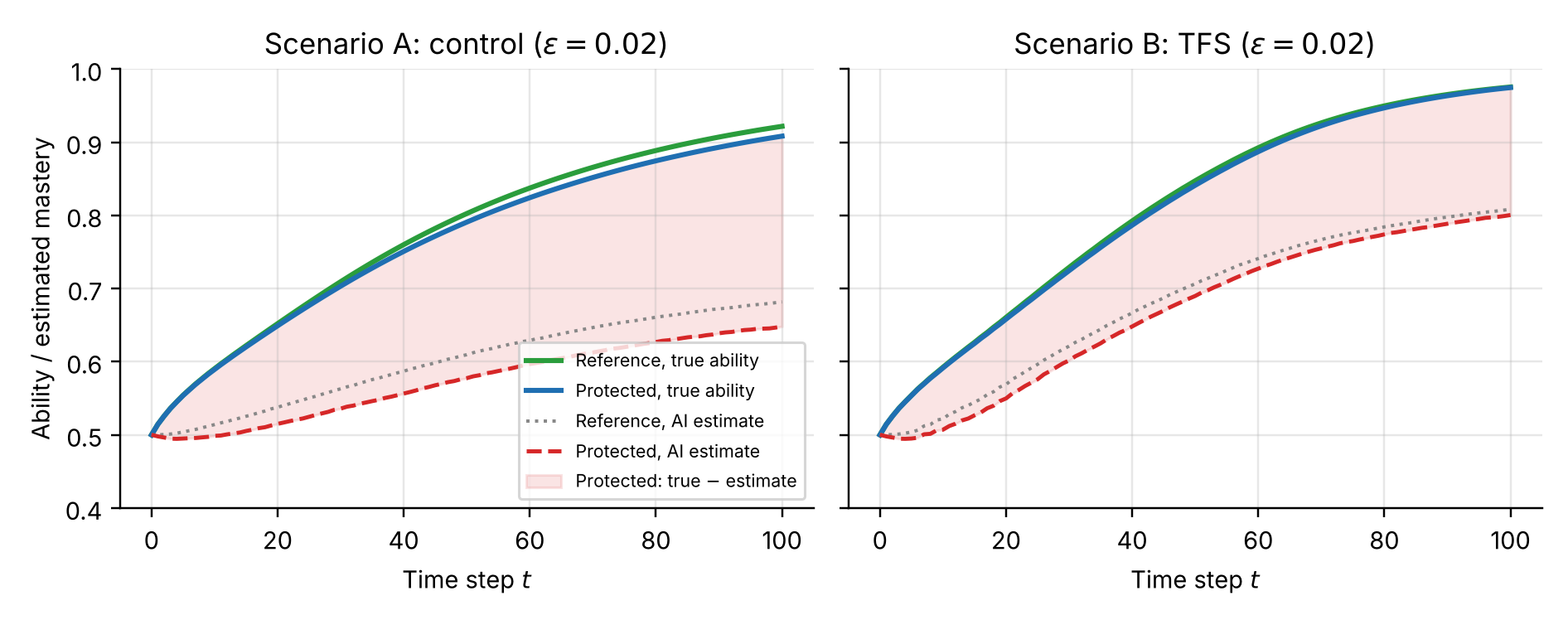} 
    \caption{Algorithmic drift persists at a lower noise parameter $\epsilon=0.02$ (online tracker, mean trajectories over $R=50$ replications; gap widening $\Delta=0.013$ under control versus $0.001$ under TFS).}
    \label{fig:drift2}
\end{figure}

To further verify the persistence of these artifacts under minimal noise conditions, we plotted the agent trajectories for a reduced noise parameter (Fig~\ref{fig:drift2}). As illustrated in Scenario A, the phenomenon of drift decouples the student's actual progress from the system's perception; while the protected group's true ability nearly converges with the reference group, the AI's estimated mastery lags, generating an area of spurious deficit beyond the lag that affects the reference group. This indicates that even when the observation noise is insufficient to stall learning, it remains sufficient to corrupt the validity of the assessment record. It creates a persistent disparity between a student's actual human capital and their digital profile. Scenario B reconfirms the efficacy of the TFS, which forces the convergence of the estimated and true states, thereby substantially reducing the drift even in these subtler accumulation scenarios.

To address the concern that our results may depend on an arbitrary choice of $\epsilon$, we performed a sensitivity analysis varying the false negative rate from $0$ to $30\%$ for both the control policy and TFS.

\begin{table}[h]
\begin{adjustwidth}{-0.5in}{-0.5in}
\centering
\footnotesize
\begin{tabular}{lcccccccc}
\toprule
 & \multicolumn{4}{c}{\textbf{Control}} & \multicolumn{3}{c}{\textbf{TFS}} \\
\cmidrule(lr){2-5}\cmidrule(lr){6-8}
\textbf{Noise $\epsilon$} & Final ability ($S_{prot}$) & $\Delta$ & $D$ ($S_{prot}$) & $G_b$ & Final ability ($S_{prot}$) & $\Delta$ & $G_b$ \\
\midrule
0.00 & $0.922$ & $0.000$ & $0.235$ & $+0.000$ & $0.975$ & $0.000$ & $+0.000$ \\
0.01 & $0.916$ & $0.006$ & $0.242$ & $-0.009$ & $0.975$ & $0.000$ & $-0.005$ \\
0.02 & $0.909$ & $0.013$ & $0.248$ & $-0.019$ & $0.975$ & $0.000$ & $-0.011$ \\
0.03 & $0.901$ & $0.020$ & $0.255$ & $-0.028$ & $0.974$ & $0.001$ & $-0.016$ \\
0.05 & $0.884$ & $0.037$ & $0.268$ & $-0.047$ & $0.972$ & $0.002$ & $-0.026$ \\
0.08 & $0.855$ & $0.067$ & $0.289$ & $-0.076$ & $0.967$ & $0.007$ & $-0.040$ \\
0.10 & $0.833$ & $0.089$ & $0.303$ & $-0.094$ & $0.962$ & $0.011$ & $-0.049$ \\
0.15 & $0.774$ & $0.147$ & $0.336$ & $-0.137$ & $0.948$ & $0.024$ & $-0.069$ \\
0.20 & $0.715$ & $0.206$ & $0.365$ & $-0.174$ & $0.929$ & $0.043$ & $-0.085$ \\
0.30 & $0.616$ & $0.306$ & $0.406$ & $-0.224$ & $0.881$ & $0.089$ & $-0.111$ \\
\bottomrule
\end{tabular}
\caption{Sensitivity of learning outcomes to observation noise ($R=25$ replications per cell; reference-group final ability $0.922$ under control and $0.975$ under TFS throughout). Gap widening $\Delta$ carries a replication-level $95\%$ bootstrap CI of at most $\pm0.001$ at every row; $G_b$ is the estimation-bias gap.}
\label{tab:sensitivity}
\end{adjustwidth}
\end{table}

As shown in Table \ref{tab:sensitivity}, algorithmic drift is nonlinear in the noise rate, and---consistent with the bifurcation in Proposition~\ref{prop:bifurcation}---the marginal harm per unit of $\epsilon$ \emph{increases} as $\epsilon$ crosses the trapping threshold. For the calibration used here the closed-form threshold of the hard-indicator idealization is $\epsilon^\star\approx0.040$ (the smooth kernel delays the collapse of the asymptotic ability to $\epsilon\approx0.082$); below it (at $\epsilon=0.01$) the gap is small ($\Delta=0.006$), reflecting a pure exposure deficit while agents remain within the ZPD, whereas above it the gap rises steeply (per-unit slope $\approx0.79$ across $1$--$5\%$ versus $\approx1.17$ across $10$--$15\%$). This accelerating profile past $\epsilon^\star$ is the simulation signature of the trapping transition: once the biased curriculum pushes the protected group's served difficulty outside the ZPD, learning decays rather than merely lagging. Two practical implications follow. First, even a highly accurate grader operating near the threshold produces a measurable, statistically reliable gap, so accuracy alone is not a sufficient fairness guarantee. Second, because the harm is concentrated around $\epsilon^\star$, small reductions in grader friction near the threshold yield disproportionately large fairness gains---an actionable target for dialect-robust grading.

\subsection{Initial State and Group Heterogeneity}
\label{sec:initial_state}
Two natural questions are how the initial state of the agents affects the results, and how the findings bear on real cohorts in which groups differ in prior knowledge and learning rate. Table~\ref{tab:initial} reports three experiments at $\epsilon=0.05$ ($R=25$). First, we varied the initial-ability distribution $\theta_i(0)\sim\mathcal{N}(\mu_0,\sigma_0)$ over $\mu_0\in\{0.3,0.5,0.7\}$ and $\sigma_0\in\{0.05,0.10,0.20\}$ for both groups. Second, we misspecified the estimator's initial belief $\hat\theta_0=\mu_0+\xi$ by $\xi\in\{-0.2,\dots,+0.2\}$, so that the tutor starts either pessimistic or optimistic about every learner. Third, we relaxed ability equivalence: the protected group was initialized with a different mean prior ability ($\mu_0^{prot}\in\{0.40,\dots,0.60\}$ against $0.50$ for the reference group) or a different mean learning rate ($\bar\alpha^{prot}\in\{0.016,\dots,0.024\}$ against $0.020$). In the third experiment the groups differ even with a noiseless channel, so we report the gap at $\epsilon=0$, the gap at $\epsilon=0.05$, and their difference, the \emph{excess gap} attributable to the channel.

\begin{table}[H]
\begin{adjustwidth}{-0.5in}{-0.5in}
\centering
\footnotesize
\begin{tabular}{lcccc}
\toprule
\multicolumn{5}{l}{\emph{Initial-ability distribution of both groups, $\theta_i(0)\sim\mathcal{N}(\mu_0,\sigma_0)$}} \\
$\mu_0$ & $\sigma_0$ & Control $\Delta$ & Control $D$ & TFS $\Delta$ \\ \midrule
0.3 & 0.05 & $+0.042$ & $0.246$ & $+0.004$ \\
0.3 & 0.10 & $+0.040$ & $0.247$ & $+0.002$ \\
0.3 & 0.20 & $+0.034$ & $0.252$ & $-0.006$ \\
0.5 & 0.05 & $+0.039$ & $0.267$ & $+0.004$ \\
0.5 & 0.10 & $+0.037$ & $0.268$ & $+0.002$ \\
0.5 & 0.20 & $+0.032$ & $0.273$ & $-0.002$ \\
0.7 & 0.05 & $+0.023$ & $0.282$ & $+0.002$ \\
0.7 & 0.10 & $+0.024$ & $0.283$ & $+0.001$ \\
0.7 & 0.20 & $+0.022$ & $0.285$ & $-0.001$ \\
\midrule \multicolumn{5}{l}{\emph{Initial estimate offset $\hat\theta_0=\mu_0+\xi$}} \\
$\xi$ & & Control $\Delta$ & Control $D$ & TFS $\Delta$ \\ \midrule
$-0.2$ & & $+0.044$ & $0.274$ & $+0.007$ \\
$-0.1$ & & $+0.041$ & $0.271$ & $+0.005$ \\
$+0.0$ & & $+0.037$ & $0.268$ & $+0.002$ \\
$+0.1$ & & $+0.034$ & $0.268$ & $-0.000$ \\
$+0.2$ & & $+0.031$ & $0.268$ & $-0.002$ \\
\midrule \multicolumn{5}{l}{\emph{Protected group with different prior ability $\mu_0^{prot}$ (reference $0.50$)}} \\
$\mu_0^{prot}$ & $\Delta_0$ & $\Delta_{0.05}$ & Excess & TFS $\Delta_{0.05}$ \\ \midrule
0.40 & $-0.059$ & $-0.022$ & $+0.037$ & $-0.074$ \\
0.45 & $-0.032$ & $+0.006$ & $+0.038$ & $-0.038$ \\
0.50 & $-0.000$ & $+0.037$ & $+0.038$ & $+0.003$ \\
0.55 & $+0.036$ & $+0.072$ & $+0.037$ & $+0.047$ \\
0.60 & $+0.075$ & $+0.110$ & $+0.035$ & $+0.093$ \\
\midrule \multicolumn{5}{l}{\emph{Protected group with different mean learning rate $\bar\alpha^{prot}$ (reference $0.020$)}} \\
$\bar\alpha^{prot}$ & $\Delta_0$ & $\Delta_{0.05}$ & Excess & TFS $\Delta_{0.05}$ \\ \midrule
0.016 & $+0.049$ & $+0.088$ & $+0.039$ & $+0.035$ \\
0.018 & $+0.021$ & $+0.059$ & $+0.039$ & $+0.015$ \\
0.020 & $-0.001$ & $+0.037$ & $+0.038$ & $+0.002$ \\
0.022 & $-0.016$ & $+0.020$ & $+0.036$ & $-0.006$ \\
0.024 & $-0.027$ & $+0.007$ & $+0.034$ & $-0.011$ \\
\bottomrule
\end{tabular}
\caption{Initial-state and group-heterogeneity sensitivity at $\epsilon=0.05$ ($R=25$). Top: initial-ability distribution of both groups (control gap $\Delta$, control drift $D$, and TFS gap). Middle: initial estimate offset $\xi$ of the tracker. Bottom: protected group with a different prior ability or learning rate; $\Delta_0$ is the gap with a noiseless channel, $\Delta_{0.05}$ the gap with the biased channel, and the excess $\Delta_{0.05}-\Delta_0$ is the drift-attributable component; the last column is the TFS gap at $\epsilon=0.05$.}
\label{tab:initial}
\end{adjustwidth}
\end{table}

Three findings emerge. First, the control gap is present for every initial-ability distribution (range $\Delta\in[0.022,0.042]$ across the nine cells) and TFS reduces it in every cell (TFS $\Delta\in[-0.006,+0.004]$); a higher starting ability compresses the gap because both groups approach the mastery ceiling within the horizon, but neither the mean nor the spread of the initial state changes the sign or the order of magnitude of the effect. Second, misspecifying the tutor's initial belief by up to $\pm0.2$ changes the control gap only marginally ($\Delta\in[0.031,0.044]$; TFS $\Delta\le0.007$), because the tracker forgets its initialization within a few tens of steps; the drift is driven by the persistent channel, not by the starting point. Third, when the groups differ in prior ability or learning rate, the noiseless gap $\Delta_0$ is non-zero and carries either sign, but the excess gap attributable to the biased channel is nearly constant at $+0.034$ to $+0.039$---essentially the ability-equivalent value $\Delta=0.038$---so the drift \emph{adds} to whatever baseline difference exists rather than being an artifact of ability equivalence, and TFS removes the excess in every case (its gap tracks $\Delta_0$ to within $0.018$). The ability-equivalence assumption of the main experiments is therefore an identification device that isolates the mechanism; it does not generate it.

\subsection{Robustness, Ablations, and Partial Failure Conditions}
\label{sec:robustness}
We evaluated the robustness of these results against the critique of simplified learner dynamics along three axes: the learning-gain rule, the estimator family, and the noise process.

\paragraph{Learning-gain rule.} We re-ran the control condition under a \textit{soft-gain} rule, where learning occurs at $25\%$ efficiency even when content falls outside the ZPD, and under a \textit{hard-indicator} rule. Under soft gain the \emph{capability} gap nearly vanishes ($\Delta=0.008$) because out-of-ZPD practice still yields slow progress, yet the \emph{epistemic} drift actually rises ($D=0.299$): while students continue to learn, the system's estimate of their mastery remains depressed ($\hat{p}\approx0.64$ against a true mean of $0.97$). The hard-indicator rule reproduces the kernel result closely ($\Delta=0.041$, $D=0.257$). The core phenomenon---corruption of the assessment record---therefore holds regardless of the strictness of the learning update; only its expression as capability versus epistemic harm shifts.

\paragraph{Estimator family.} Proposition~\ref{prop:bifurcation} indicates that drift is a property of the policy--observation feedback loop rather than of any particular knowledge-tracing model. In addition to the LSTM-DKT experiment of Section~\ref{sec:lstm_results}, we replaced the online tracker with a structurally different moment-matching estimator (an EMA of the observed success rate inverted through the IRT link): the gap and drift again rise monotonically with $\epsilon$ ($\Delta$ from $0.004$ at $\epsilon=0.01$ to $0.077$ at $\epsilon=0.15$; $D$ from $0.36$ to $0.41$). Varying the tracker's step size $\eta\in\{0.1,0.2,0.4\}$ changes the control gap ($\Delta=0.056$, $0.037$, $0.023$) but TFS closes it at every setting ($\Delta\le0.004$).

\paragraph{Noise process.} The baseline channel is a memoryless flip with constant rate. We re-ran control and TFS with (i) a \emph{difficulty-dependent} channel whose flip probability rises linearly with item difficulty (mean $\epsilon$), (ii) a \emph{bursty} two-state Markov channel in which friction switches on and off with persistence $0.9$ (mean $\epsilon$), and (iii) a \emph{continuous} per-learner friction rate $\epsilon_i\sim\mathrm{Beta}$ with mean $\epsilon$ and no group labels, with the ``protected group'' defined post hoc as the most-affected $20\%$. At mean $\epsilon=0.05$ the control gap is $\Delta=0.031$ (difficulty-dependent) and $0.035$ (bursty), against $0.038$ for the memoryless channel, and TFS reduces these to $0.002$ and $0.007$; under the continuous channel the most-affected $20\%$ carry a mean friction rate of about $0.16$ rather than $0.05$, and their gap is correspondingly larger ($\Delta=0.137$ under control, $0.033$ under TFS). Under the continuous channel the realized learning gain decreases monotonically with the learner's individual friction rate, so the harm is a dose--response relation at the individual level rather than a group label.

Furthermore, we explored boundary conditions under extreme noise. At $\epsilon = 0.20$ the unmitigated gap is large ($\Delta = 0.206$); TFS reduces it substantially but no longer to zero, leaving a residual gap of $\Delta = 0.043$, and when this is combined with a depressed learning-rate population (mean $\alpha_i$ lowered to the low-quartile regime) the residual persists at a comparable level ($\Delta = 0.050$); at $\epsilon=0.30$ the residual is $0.090$ against an unmitigated $0.306$. Thus TFS degrades gracefully rather than collapsing: as anticipated by Corollary~\ref{cor:tfs} and condition (C1), the guarantee weakens once the observation channel is corrupted enough that probes themselves frequently register false negatives, so a residual gap survives. TFS is therefore a targeted regularizer, not a panacea: it breaks the standard feedback loop but cannot fully substitute for reducing $\epsilon$ at its source (e.g., dialect-robust grading).

\paragraph{Hyperparameter ablation.} TFS exposes two hyperparameters: the patience $\tau$ (consecutive pessimism steps before a probe) and the cohort margin $m$ that defines how far below the cohort's typical estimated level a learner must fall to be counted as ``behind.'' We ablate both over $m\in\{0.05,0.10,0.15\}$ and $\tau\in\{3,5,8,12\}$ at $\epsilon=0.05$ ($R=25$); Table~\ref{tab:ablation} reports the residual gap $\Delta$. Across the entire grid the residual gap stays within $[-0.006, +0.026]$ of zero, confirming that the near-elimination of the gap is not knife-edge in the hyperparameters. The sign structure matches the theory: at small $\tau$ the system probes eagerly and slightly \emph{over}-corrects (negative residual, i.e., the protected group marginally exceeds reference growth), whereas at large $\tau$ it tolerates longer remedial loops and a small positive residual survives. Replacing the cohort-relative trigger by the absolute trigger ($\beta<\mu_{threshold}$) yields $\Delta=0.025$ at $\epsilon=0.05$ and $0.098$ at $\epsilon=0.15$ (cohort trigger: $0.003$ and $0.024$), so the absolute variant remains usable when no cohort reference exists (condition C2), at the cost of a larger residual.

\begin{table}[h]
\centering
\begin{tabular}{lcccc}
\hline
\textbf{Cohort margin $m$} & $\tau=3$ & $\tau=5$ & $\tau=8$ & $\tau=12$ \\ \hline
0.05 & $-0.006$ & $+0.000$ & $+0.013$ & $+0.023$ \\
0.10 & $-0.005$ & $+0.002$ & $+0.015$ & $+0.025$ \\
0.15 & $-0.001$ & $+0.007$ & $+0.018$ & $+0.026$ \\ \hline
\end{tabular}
\caption{TFS hyperparameter ablation at $\epsilon=0.05$: residual protected-group gap $\Delta$ as a function of patience $\tau$ and cohort margin $m$ ($R=25$). The recovered gap is near zero across the grid.}
\label{tab:ablation}
\end{table}

\subsection{Failure Conditions of TFS: Probe Side Effects}
\label{sec:probefail}
Condition (C3) of Section~\ref{sec:tfs} assumes that a failed probe is harmless. We relaxed it in two ways. A \emph{frustration} penalty multiplies the learner's learning rate by $(1-\phi)$ for the five steps following a failed probe, $\phi\in\{0,0.25,0.5,1\}$ ($\phi=1$: no learning at all for five steps after every failed probe). An \emph{entrenchment} penalty lowers the tutor's estimate by $\psi\in\{0,0.05,0.10\}$ after a failed probe, modeling a system that treats a failed probe as evidence of deficit. Because a probe is placed near the learner's estimated level and graded through the same channel, roughly half of all probes fail ($0.47$ at $\epsilon=0.05$), so both penalties are exercised frequently. Table~\ref{tab:probefail} reports the residual gap at $\epsilon=0.05$ and $0.15$ ($R=25$).

\begin{table}[H]
\begin{adjustwidth}{-0.5in}{-0.5in}
\centering
\footnotesize
\begin{tabular}{lccc}
\toprule
\multicolumn{4}{l}{\emph{$\epsilon=0.05$}} \\
$\phi$ \textbackslash\ $\psi$ & $0.0$ & $0.05$ & $0.1$ \\ \midrule
0.00 & $+0.002$ ($0.972$) & $+0.004$ ($0.967$) & $+0.004$ ($0.964$) \\
0.25 & $+0.003$ ($0.969$) & $+0.004$ ($0.965$) & $+0.005$ ($0.962$) \\
0.50 & $+0.003$ ($0.965$) & $+0.005$ ($0.961$) & $+0.005$ ($0.958$) \\
1.00 & $+0.006$ ($0.952$) & $+0.007$ ($0.949$) & $+0.008$ ($0.947$) \\
\midrule
\multicolumn{4}{l}{\emph{$\epsilon=0.15$}} \\
$\phi$ \textbackslash\ $\psi$ & $0.0$ & $0.05$ & $0.1$ \\ \midrule
0.00 & $+0.024$ ($0.948$) & $+0.027$ ($0.942$) & $+0.029$ ($0.938$) \\
0.25 & $+0.026$ ($0.944$) & $+0.029$ ($0.938$) & $+0.031$ ($0.934$) \\
0.50 & $+0.029$ ($0.939$) & $+0.032$ ($0.933$) & $+0.033$ ($0.929$) \\
1.00 & $+0.038$ ($0.920$) & $+0.041$ ($0.915$) & $+0.043$ ($0.912$) \\
\bottomrule
\end{tabular}
\caption{TFS under probe side effects ($R=25$): residual gap $\Delta$ (and protected-group final ability in parentheses) as a function of the frustration penalty $\phi$ and the entrenchment penalty $\psi$. The unmitigated control gap is $0.037$ at $\epsilon=0.05$ and $0.147$ at $\epsilon=0.15$.}
\label{tab:probefail}
\end{adjustwidth}
\end{table}

Both side effects erode the benefit monotonically, and the erosion is larger at higher noise, where more probes fail. In the worst cell---no learning for five steps after every failed probe and a $0.10$ downward revision of the estimate---the residual gap rises to $+0.008$ at $\epsilon=0.05$ and $+0.043$ at $\epsilon=0.15$, against unmitigated gaps of $0.037$ and $0.147$; TFS therefore still removes $79\%$ and $71\%$ of the gap under the most adverse assumptions we examined, but the protected group's final ability falls from $0.972$ to $0.947$, so the cost of probing is borne by the learners who are probed. Two design implications follow: probes should be framed and paced to minimize frustration, and a failed probe must not be allowed to lower the estimate below its pre-probe level, since entrenchment converts the diagnostic into an additional source of pessimism.

\subsection{Statistical Analysis}
\label{sec:stats}
The first submission reported a Welch $t$-test on per-agent final abilities and a Cohen's $d$ of $42.98$ in the baseline comparison table. The latter was computed on replication means: because the standard deviation of a mean over roughly $2{,}000$ protected learners is of the order of $0.003$, any difference of practical size yields a standardized value in the tens, and the number is not comparable to conventional effect-size benchmarks. We now report the contrast on the two scales explicitly. At the replication level, pairing the $R=50$ seeds, the protected group's final ability under TFS exceeds that under the control policy by $+0.088$ ($95\%$ bootstrap CI $[+0.087, +0.088]$; paired $t(49)=273.0$, $p=1.2 \times 10^{-79}$; Wilcoxon $p=1.8 \times 10^{-15}$), with the difference positive in $100\%$ of replications; the corresponding gap widening changes by $-0.035$ ($95\%$ CI $[-0.035, -0.034]$). At the learner level, pooling all protected learners over replications ($n=99,509$ per policy), the standardized effect size is Cohen's $d=0.83$ (within-replication $d$: mean $0.83$, SD $0.03$ across seeds), a large effect relative to between-learner variability; the learner-level $d$ values in Table~\ref{tab:baselines} are computed in the same way. We stress that with $N$ in the thousands, learner-level significance is guaranteed for any nonzero effect; the replication-level interval and the learner-level effect size are the substantive quantities, and both indicate that the recovery attributable to regularization is substantial rather than merely detectable.
\section{Discussion}

The findings of this study have notable implications for the long-term deployment of AI-integrated higher education. While current discourse often frames educational technology sustainability in terms of computational energy efficiency, our results highlight a parallel dimension of \textit{human capital} sustainability. An educational system that systematically under-skills a demographic segment due to algorithmic artifacts misallocates instructional effort and risks perpetuating systemic inequities that challenge the objectives of UN Sustainable Development Goal 4 (SDG 4). Before drawing implications we state precisely what the evidence establishes.

\subsection{What the Evidence Establishes, and What Remains to be Validated}
\label{sec:evidence}
The paper's claims rest on three kinds of evidence with different epistemic status; we keep them separate and state what remains open.
\begin{itemize}[leftmargin=1.4em, itemsep=2pt, topsep=2pt]
    \item \emph{Established analytically} (Proposition~\ref{prop:bifurcation}, Corollary~\ref{cor:tfs}): in a reduced mean-field model with a hard ZPD indicator, a policy that targets a fixed observed success rate through an asymmetric false-negative channel serves the affected group easier content for every $\epsilon>0$, with a closed-form deficit, and traps it above a computable threshold $\epsilon^\star$; periodic ZPD re-entry prevents unbounded decay. These statements are theorems about the model and hold for any estimator consistent for the observed signal.
    \item \emph{Demonstrated in simulation} (Sections~\ref{sec:baseline_results}--\ref{sec:probefail}): with heterogeneous synthetic learners, a smooth ZPD kernel, and either the online tracker or an LSTM-DKT in the loop, the estimation bias, the exposure gap, the ability gap, and the accelerating harm past the trapping threshold appear as predicted. The threshold lies at larger $\epsilon$ for the better-calibrated LSTM. TFS brings the contrasts near parity where $\epsilon$-greedy, UCB and Thompson sampling do not. The qualitative findings survive changes to the learning rule, estimator, noise process, initial state, group heterogeneity, and TFS hyperparameters, and degrade gracefully under extreme noise and probe side effects. The absolute magnitudes depend on the parameters of Table~\ref{tab:params}.
    \item \emph{Measured} (Section~\ref{sec:llm_results}): current open-weight graders exhibit the asymmetric false-negative channel on a matched-pair bank---a $5$-point false-negative asymmetry for the most affected grader, coinciding with the simulation baseline, and a consistent downward shift of grader confidence for second-language answers across graders---with no false negatives on Standard-English answers. This is a pilot measurement of the channel's existence and order of magnitude, not a population estimate.
    \item \emph{Not yet validated}: that real learners' ability responds to served difficulty as Eq.~\eqref{eq:learn} assumes; that the closed loop operates in a deployed platform with its instructors, peers, and multimodal assessments; that the measured $\epsilon$ transfers across dialects, subjects, and response lengths; and that TFS probes are tolerated by real students. The real-world implications below are conditional on these points.
\end{itemize}

\subsection{The Dynamics of Algorithmic Drift}
The simulation provides a controlled demonstration of how generative-adaptive systems might amplify initial assessment errors into structural deficits. The phenomenon of \textit{algorithmic drift} illustrates that fairness is not a static property of a model but a dynamic state of a coupled system. The feedback loop observed in Scenario A functions as a recursive degradation of the learner profile. When the AI system misinterpreted a student's linguistic friction as a conceptual failure, it adapted by lowering the complexity of the generated content.

Our sensitivity analyses indicate that this phenomenon persists when adjusting for heterogeneous learning speeds, softer gain rules, different initial states, and group differences in prior ability. The system's \textit{perception} of the student remains artificially depressed, creating a dynamic where the model validates its own pessimism. Paradoxically, the student's success on easier, remedial tasks reinforces the model's biased prediction, demonstrating how highly accurate short-term predictions can lead to long-term trajectory decoupling.

Our contribution relative to prior work on feedback loops---repeated risk minimization \cite{hashimoto2018fairness}, performative prediction, and the Matthew effect in learning analytics \cite{yu2023should,liu2022equity}---is twofold. Prior work establishes that error can compound; we specify \emph{a mechanism particular to generative-adaptive tutoring} (an adaptive difficulty policy coupled to an asymmetric grader) and, crucially, characterize its phase structure. Proposition~\ref{prop:bifurcation} shows the harm is not a smooth accumulation but a bifurcation: a benign ``under-challenged but progressing'' regime below a computable threshold $\epsilon^\star$, and a ``trapped and declining'' regime above it. This distinction is practically consequential, because it implies that an intervention must guarantee periodic ZPD re-entry (Corollary~\ref{cor:tfs}) rather than merely reduce average error---explaining why generic exploration is theoretically insufficient and motivating the loop-duration trigger of TFS.

\subsection{Quantifying Instructional Misallocation}
From a systems engineering perspective, unmitigated algorithmic drift represents a significant inefficiency. The modeled discrepancy between the protected group's true ability ($\theta=0.884$) and the AI's estimate ($\hat{p}=0.593$) at $\epsilon=0.05$, over and above the estimator lag that the reference group also experiences, signifies that the system was expending instructional resources to generate remedial content for students capable of more advanced work (remedial items on $40\%$ of protected-group steps against $29\%$ for the reference group)---a misallocation that grows as $\epsilon$ increases.

This constitutes a misallocation of time, energy, and digital infrastructure to pedagogical interventions that yield suboptimal outcomes. While the near-complete recovery achieved by the TFS in Scenario B indicates the potential, within the model, to reclaim this lost capability, it primarily serves to illustrate that sustainable AI must be designed with explicit resilience against asymmetric observation noise.

\subsection{Limitations and External Validity}
The study has four main limitations. First, the learners are synthetic. The observation-noise parameter has now been measured in current graders (Section~\ref{sec:llm_results}), but the learning trajectories are simulated in a POMDP environment that excludes instructors, peers, and multimodal assessment. The effect sizes of the control scenario ($\Delta=0.038$ at $\epsilon=0.05$, $0.147$ at $\epsilon=0.15$) therefore describe the autonomous loop in isolation, and their magnitude depends on the coupling parameters of Table~\ref{tab:params}; the robust findings are the qualitative bifurcation, the accelerating harm past $\epsilon^\star$, and the separation of TFS from generic exploration. Second, the observation channel is a stylized construct. It compresses linguistic friction into a scalar false-negative rate, treats the protected group as homogeneous, and ignores intersectionality and false positives; the difficulty-dependent, bursty, and continuous per-learner channels of Section~\ref{sec:robustness} relax but do not remove this simplification. Third, the model does not represent modality preferences such as visual, auditory, or hands-on learning. We did not add them because the empirical literature finds no reliable interaction between such preferences and instructional modality \cite{pashler2008learning}; learner heterogeneity enters instead through the per-learner friction rate $\epsilon_i$ and learning rate $\alpha_i$, and a modality-aware content generator would add state to the policy without altering the estimator--policy loop. Fourth, TFS is strikingly effective because a passed probe immediately reveals the learner's demonstrated level and a failed probe is costless. Conditions (C1)--(C4) of Section~\ref{sec:tfs} state when this can be expected to hold, and Section~\ref{sec:probefail} shows that frustration and entrenchment after failed probes erode but do not eliminate the benefit; probe frequency, placement, and framing would need to be designed with these side effects in mind. Future work should validate the mechanism by mapping linguistic markers to longitudinal curriculum velocity in anonymized logs from deployed adaptive platforms (such as EdNet or ASSISTments) and by measuring $\epsilon$ on independently annotated banks that span dialects, subjects, and response lengths.

\subsection{Policy Implications for Institutional Adaptation}

The recommendations in this subsection are conditional on the mechanism operating in deployed systems as it does in the model (Section~\ref{sec:evidence}); they describe what institutions could do if it does, not what the present evidence shows they must do. The finding that standard optimization in GenAI can construct feedback loops of inequality suggests a need to update how higher education institutions govern EdTech. Current procurement frameworks predominantly rely on \textit{static auditing}—evaluating a model’s training data or decision boundaries prior to deployment. Our theoretical framework shows that a model fair at initialization ($t=0$) can degenerate into biased behavior through sustained interaction ($t=n$).

To support the equitable deployment of AI-integrated learning environments, we propose a multi-layered framework for institutional adaptation:

\paragraph{From static certification to longitudinal auditing:}
Institutions should consider requiring vendors to provide evidence of \textit{temporal stability} rather than relying solely on static accuracy metrics. Requests for proposals for adaptive learning systems could mandate longitudinal impact statements that simulate the trajectory of diverse learner profiles over extended periods, utilizing dynamic metrics such as the \textit{drift divergence} defined in this study.

\paragraph{Implementing fairness-aware exploration frameworks:}
Unlike generic RL exploration such as $\epsilon$-greedy methods which optimize for average global reward, adaptive educational systems require targeted fairness corrections for asymmetric demographic noise. Institutions should advocate for technical standards that require generative-adaptive systems to include ``pessimism breakers'' similar to our TFS. These mechanisms help ensure that learners are not permanently relegated to a remedial track without periodic opportunities to demonstrate latent capability.

\paragraph{The instructor as the drift detector:}
As AI assumes a larger role in content generation, the human instructor's role must evolve toward algorithmic oversight. Learning management systems should be redesigned to visualize potential algorithmic drift. Instead of simply reporting a low mastery score, dashboards could flag instances where a student’s performance history conflicts with the AI’s predictive confidence, enabling instructors to intervene precisely when human-in-the-loop context is needed to correct a system's spurious deficit \cite{Kucuk2025}.

\paragraph{Centralized trajectory monitoring:}
To align with UN SDG 4, institutions may establish centralized oversight bodies to monitor aggregate learning trajectories. By analyzing anonymized progression data, universities can detect if specific demographic groups systematically experience lower curriculum velocity despite passing grades. Such patterns would indicate that the adaptive infrastructure might be optimizing for engagement or easy completions rather than cognitive growth, signaling a need for systemic recalibration.

By embedding these regenerative mechanisms and governance protocols into the institutional architecture, higher education can better ensure that the adoption of GenAI supports the long-term sustainability of the learning ecosystem, fostering an environment where technological efficiency computationally supports equitable student outcomes.

\subsection{Human-Centred and Institutional Implementation}

The intervention proposed here addresses algorithmic drift at the level of the adaptive policy, but sustaining fairness and educational effectiveness over time depends on more than algorithmic optimization. Higher education has absorbed successive waves of technology, from personal computing to online learning, without displacing its core purpose of developing knowledge, critical thinking, and graduate capability. Adaptive AI is best understood in the same terms: a tool that augments rather than replaces educator expertise and professional judgment. Realizing this in practice requires institutional investment in people alongside technical mechanisms such as the TFS. Strengthening AI literacy, supporting faculty development, and reconsidering assessment design keep human oversight available to interpret AI-generated recommendations, to weigh contextual factors that the model does not observe, and to keep educational decisions aligned with institutional values and learner needs, consistent with emerging work on responsible AI implementation in higher education \cite{chan2023, porayska2024ethics}.

The phenomenon also carries a reflexive institutional implication. If adaptive systems require continuous monitoring to remain fair, the institutions that deploy them plausibly require a similarly ongoing approach to evaluating their own AI-enabled practices, treating adaptive AI as part of a continuous cycle of institutional learning rather than a one-time implementation. The longitudinal data these systems generate can complement existing evidence on teaching, curriculum effectiveness, assessment, and student progression, supporting more evidence-informed institutional decision-making \cite{holstein2022equity}. Such evidence should complement rather than replace professional academic judgment and engagement with students and staff. Framed this way, the significance of algorithmic drift extends beyond its computational statement: the same principle of continuous adaptation that stabilizes the learner--tutor dyad may encourage institutions to adopt more reflective, evidence-informed approaches to implementing and evaluating AI, contributing to its responsible and sustainable adoption in higher education \cite{williamson2020historical}.


\section{Conclusion}
This study elucidates a potential vulnerability in the deployment of GenAI for education: the tendency of adaptive systems to automate the stratification of learners through feedback loops of {algorithmic drift}. By moving beyond static bias audits and modeling the longitudinal dynamics of the learner--tutor dyad, we showed analytically and in simulation that algorithms optimizing for the learner can transform minor observation noise such as linguistic friction into systemic deficits, and we measured the triggering asymmetry in current foundation-model graders. Within the modeled assumptions, the results support the conclusion that without active regularization, GenAI tutors risk creating self-fulfilling prophecies that artificially cap the human capital development of marginalized populations; whether and how strongly the loop operates in deployed platforms remains to be established with real learner data.
However, our findings also offer a blueprint for resilience. The success of the TFS in simulation indicates that these feedback loops are not inevitable. By embedding rights to exploration into the algorithmic architecture and ensuring that epistemic uncertainty triggers a re-verification of learner capability rather than a retreat to remedial content, we can decouple observation noise from curriculum generation. As higher education transitions toward the segment of one, the definition of algorithmic fairness must evolve from a static checkbox at deployment to a continuous, regenerative process. Ultimately, realizing the promise of SDG 4 in the age of AI requires systems that are not only computationally efficient but engineered to be robust guardians of student potential over the long range.


\end{document}